\documentclass[lettersize,journal]{IEEEtran}
\usepackage{amsmath,amsfonts}
\usepackage{algorithmic}
\usepackage{algorithm}
\usepackage{array}
\usepackage{textcomp}
\usepackage{stfloats}
\usepackage{url}
\usepackage{verbatim}
\usepackage{graphicx}
\usepackage{cite}
\usepackage{enumitem}
\usepackage{hyperref}
\usepackage[capitalise,noabbrev]{cleveref}
\usepackage{authblk}

\newlist{arrowlist}{itemize}{1}
\setlist[arrowlist]{label=$\Rightarrow$}

\usepackage{graphicx}
\usepackage{xspace}
\usepackage{siunitx}
\usepackage[capitalise,noabbrev]{cleveref}
\usepackage{booktabs}
\usepackage[dvipsnames]{xcolor,colortbl}
\usepackage{tabularx}
\usepackage{subcaption}

\usepackage{amsbsy}

\usepackage[normalem]{ulem}

\newcommand{\RomanNumeralCaps}[1]
    {\MakeUppercase{\romannumeral #1}}

\makeatletter
\DeclareRobustCommand\onedot{\futurelet\@let@token\@onedot}
\def\@onedot{\ifx\@let@token.\else.\null\fi\xspace}
\DeclareRobustCommand\nodot{\futurelet\@let@token\@nodot}
\def\@nodot{\ifx\@let@token.\else~\null\fi\xspace}

\newfont{\eaddfnt}{phvr8t at 12pt}

\def\eg{\emph{e.g}\onedot} 

\def\ie{\emph{i.e}\onedot} 
 
\def\etc{\emph{etc}\onedot}

\def\etal{\emph{et al}\onedot}

\usepackage{titlesec}

\titleformat{\paragraph}
{\normalfont\normalsize\bfseries}{\theparagraph}{1em}{}
\titlespacing*{\paragraph}
{0pt}{3.25ex plus 1ex minus .2ex}{1.5ex plus .2ex}

\begin{document}

\title{ROI-NeRFs: Hi-Fi Visualization \\ of Objects of Interest within a Scene \\by NeRFs Composition}


\author[1,2]{Quoc-Anh Bui}
\author[1]{Gilles Rougeron}
\author[2]{Géraldine Morin}
\author[2]{Simone Gasparini}
\affil[1]{Université Paris-Saclay, CEA, List, F-91120, Palaiseau, France}
\affil[2]{Université de Toulouse, Toulouse INP -- IRIT, France}
\renewcommand\Affilfont{\itshape\small}


\IEEEpubid{0000--0000/00\$00.00~\copyright~2021 IEEE}

\maketitle

\begin{abstract}

Efficient and accurate 3D reconstruction is essential for applications in  cultural heritage. This study addresses the challenge of visualizing objects within large-scale scenes at a high level of detail (LOD) using Neural Radiance Fields (NeRFs). The aim is to improve the visual fidelity of chosen objects while maintaining the efficiency of the computations by focusing on details only for  relevant content.
The proposed ROI-NeRFs framework divides the scene into a Scene NeRF, which represents the overall scene at moderate detail, and multiple ROI NeRFs that focus on user-defined objects of interest. An object-focused camera selection module automatically groups relevant cameras for each NeRF training  during the \textit{decomposition} phase. In the \textit{composition} phase, a Ray-level Compositional Rendering technique combines information from the Scene NeRF and ROI NeRFs, allowing simultaneous multi-object rendering composition. Quantitative and qualitative experiments conducted on two real-world datasets, including one on a complex eighteen's century cultural heritage room, demonstrate superior performance compared to baseline methods, improving LOD for object regions, minimizing artifacts, and without significantly increasing inference time. 
\end{abstract}


\section{Introduction}
\label{sect:intro}
In the era of Industry $4.0$, efficient reconstruction of real-world scenes plays a critical role in advancing computers' understanding of 3D environments. This technology has broad applications across various domains, particularly in industrial settings and cultural heritage. For large-scale industrial scenes, accurate 3D reconstructions are essential for tasks such as facility management, quality control, and predictive maintenance.
In this paper, we focus on the cultural heritage domain, where 3D scene reconstruction has significant value. 
A key downstream application is the creation of 3D digital twins of cultural heritage sites and artifacts.
These digital twins have potential use cases in diverse domains such as archiving and conservation, where they can preserve the physical state of artifacts and historical sites for future generations~\cite{Kong2023}. 
In archaeology, 3D digital twins allow for virtual exploration, analysis, and replication of artifacts and excavation sites, aiding research and education by providing wider access to historical assets while avoiding the risk of physical damage of the original model~\cite{Haibt2024}. 
Furthermore, museums can utilize digital twins to create interactive exhibitions that provide enhanced educational experiences for visitors~\cite{Liu2024}.

With the advancements and development of scene acquisition technology, photogrammetry and scanning have become fundamental approaches for 3D reconstruction, providing reliable and efficient results. 
The classic photogrammetry pipeline employs structure from motion (SfM)~\cite{Hartley2004} and multi-view stereo (MVS) algorithms~\cite{Furukawa2015} to extract and match texture features across multiple images, subsequently estimating the dense representation of 3D models.
Depth scanning techniques, such as LiDAR, utilize laser light to measure distances to target surfaces, generating a set of non-connected 3D points.

\IEEEpubidadjcol

In recent years, Neural Radiance Field (\textbf{NeRF}) \cite{mildenhall2021nerf} has rapidly emerged.
It is a novel view synthesis (\textbf{NVS}) method, whose model is trained from a set of input scene images with known poses.
NeRF’s straightforward principle, coupled with the unprecedented quality of the generated images, has driven many new contributions and developments in the field. 
Compared to an explicit representation like in 3DGS~\cite{kerbl3Dgaussians}, NeRF offers a more compact memory usage. 
However, most NeRF-like methods use all the images of a scene to train a single NeRF that represents the entire scene. 
As a result, the inferred image of a camera viewpoint supports only one level of detail (\textbf{LOD}) throughout the entire image.
Providing a unique resolution level may become problematic in large-scale scenes, as it might lead to poor-quality images when generating views close to specific objects or regions. 
Even though observation views provided more details, the single-LOD constraint could prevent the model from effectively representing such details. 
Consequently, a region of interest (\textbf{ROI}) with finer details acquired at a higher resolution could still appear at lower LOD.
On the other hand, achieving high levels of detail across the entire large scene is challenging without incurring prohibitive computation times or memory requirements.

Works such as Mip-NeRF~\cite{barron2021mipnerf}, RING-NeRF~\cite{petit2024ring}, and VR-NeRF~\cite{VRNeRF} have applied LOD-focused methods to NeRF, enabling the model to understand the concept of scale and thereby improving novel view quality.
Others, like Block-NeRF~\cite{tancik2022blocknerf} and Mega-NeRF~\cite{Turki_2022_CVPR}, suggest partitioning the space and training multiple NeRFs in 3D sub-spaces to represent large outdoor scenes. 
However, these approaches based on space partitioning lack consideration of the actual content of the scene and, in particular, the presence of regions or objects of interest within the scene. 
ObjectNeRF~\cite{yang2021objectnerf} and UDC-NeRF~\cite{wang2023udcnerf} are approaches that focus on internal objects but require prior segmentation of the objects of interest in the input images.

This paper addresses the challenge of modeling a potentially large-scale area containing regions of interest that a user might inspect more closely and for which more, or higher resolution data, is available. 
In the cultural heritage domain, such regions include \eg statues or architectural elements of interest, often found inside churches or cathedrals, captured with numerous close-up images.
Our method relies on separating NeRFs into two groups: a scene group consisting of one or more NeRFs representing the entire scene at low to moderate LOD, and an ROIs group with a NeRF per object of interest, capable of modeling finer local details in close-up views.
We introduce \textbf{ROI-NeRFs}, a framework for decomposing and composing these groups to enable high-fidelity inference.
It features an Object-Focused Camera Selection module for automatically selecting suitable viewpoints to train individual ROI networks and a crucial Ray-Level Compositional Rendering technique for efficiently processing and filtering inference information from multiple NeRFs.

In the following, we first review prior neural rendering approaches according to their applications and improvements in \cref{sect:relatedwork}. Then, we describe our concepts and the implementation of training the two NeRF groups and combining their inference to obtain novel views in \cref{sect:method}. 
Subsequently, we present our experiments on real-world datasets in \cref{sect:exp}.
Finally, the limitations of our method and perspectives for improvements are discussed in \cref{sect:limitation} and \cref{sect:conclusion} concludes the paper.

\section{Related Work}
\label{sect:relatedwork}
One of the challenges with most learning-based methods is that they require supervision and ground truth model information. 
This information is difficult to obtain accurately from real-world scenes and is often based on data from synthetic scenes. 
Drawing on the continuity of implicit representations\cite{Park2019DeepSDFLC, Mescheder2018OccupancyNL, Chen2018LearningIF} and differentiable volumetric rendering (\textbf{DVR})~\cite{Lombardi:2019, Sitzmann2018DeepVoxelsLP, Niemeyer2019DifferentiableVR, Sitzmann2019SceneRN}, end-to-end training of a 3D model directly from 2D images has been developed. 
This advancement has led to the emergence of implicit neural representations, especially Volumetric Radiance Field-based methods, as a powerful alternative to 3D scene reconstruction, particularly in novel view synthesis (\textbf{NVS}).
Neural Radiance Field (\textbf{NeRF})~\cite{mildenhall2021nerf} introduced by Mildenhall \etal has been the seminal work for many high-quality NVS follow-up methods.
NeRF learns complex scenes as continuous particle fields of volume density and view-dependent color modeled by an MLP. Using positional encoding and DVR techniques at inference achieves unprecedented photorealistic rendering quality.
However, due to dense queries in a large MLP, training and rendering times remain time-consuming.

The NeRF method takes as input a set of calibrated images with known poses: the extrinsic parameters (camera position and orientation), and the intrinsic ones (typically, the focal length and the principal point), usually obtained via SfM tools, \eg, COLMAP~\cite{schoenberger2016sfm}.
NeRF generates images using a differentiable volume renderer, where camera rays traverse a finite cubic volume containing the scene's contents.

More specifically, the 3D position of a sampled point $\textbf{x} = (x,y,z) = \textbf{r}(t)$ and the viewing direction $\textbf{d} = (\theta,\phi)$ along a camera ray $\textbf{r}(t) = \textbf{o} + t\textbf{d}$, emitted from the camera's projection center $\textbf{o}$, are input into an MLP with weights $\boldsymbol\Theta$. The network outputs the volume density $\boldsymbol\sigma$ at that location, along with the color $\textbf{c} = (r, g, b)$ corresponding to the particle's radiance:
$$
\boldsymbol\sigma(t), \textbf{c}(t)=\operatorname{MLP}_{\Theta}(\textbf{r}(t), \textbf{d})
$$
\\
The predicted color $\hat{\mathbf{C}}(\mathbf{r})$ of the pixel is calculated by using these estimated densities and colors to approximate the volume rendering integral through numerical quadrature, as discussed by Max \cite{Max1995OpticalMF}:
$$
\begin{gathered}
\hat{\mathbf{C}}(\mathbf{r})=\sum_k T\left(t_k\right) \alpha\left(\boldsymbol\sigma\left(t_k\right) \delta_k\right) \textbf{c}\left(t_k\right), \\
T\left(t_k\right)=\exp \left(-\sum_{k^{\prime}=1}^{k-1} \boldsymbol\sigma\left(t_{k^{\prime}}\right) \delta_{k^{\prime}}\right), \quad \alpha(x)=1-\exp (-x),
\end{gathered}
$$
where $T\left(t_k\right)$ denotes the accumulated transmittance, accounting for occlusion along the ray, and $ \delta_k = t_{k+1} - t_k $ represents the distance between two adjacent points along the ray.

During training, the squared errors measured from the pixels of input images are back-propagated through differentiable volume renderer to update the MLP weights. 
As a result, NeRF MLP learns a latent representation of the geometry and directional appearance of the scene. 
At inference, novel views from unknown camera poses are generated from the trained MLP using the same volume rendering technique.
Since the model size and the number of samples per ray are fixed before training, NeRF relies solely on the distribution of points from the coarse-to-fine volume sampling strategy to retrieve visible content.
As a result, NeRF supports only a single LOD, providing uniform detail across the entire image without clearly distinguishing between content near the camera and distant objects.

\textbf{Level of detail - High fidelity.} Level of detail (LOD) concepts were first applied to neural representations in Mip-NeRF~\cite{barron2021mipnerf} and VolSDF~\cite{takikawa2021nglod}. Similar to image processing and rendering, this helps adjust the complexity of the content in the scene, thereby reducing the aliasing effect.
Mip-NeRF, in particular, used integrated position encoding (IPE) over conical frustums of pixel-footprint-cone-casting, to encode scene representation at various scales.
Instant-NGP (iNGP)~\cite{mueller2022instant} adopted a hash encoding approach using a pyramid of multi-scale 3D grids to store learned features in a hash table, which were then processed by a compact MLP. This structure enables faster computation during both training and inference stages and allows control over the entire scene's LOD by adjusting two parameters: $N_{max}$, the finest resolution per axis of the highest grid level, and $T$, the fixed size of the GPU hash tables containing the features.
Developed on top of both Mip-NeRF and iNGP, Zip-NeRF~\cite{barron2023zipnerf} used supersampling and reweighting to parameterize frustum along each ray using iNGP’s multiresolution grid, allowing to achieve anti-aliasing and scale-reasoning while being significantly faster.
BungeeNeRF~\cite{xiangli2022bungeenerf} progressively trained NeRF MLP blocks to render satellite-ground cityscape scenes with extreme variations in scale. Each block has its own output, allowing the model to reasonably learn the finer details in the later blocks.
RING-NeRF~\cite{petit2024ring} represented 3D scenes as a continuous multi-scale model with an invariant decoder across spatial and scale domains. 
Its continuous coarse-to-fine optimization further enhanced reconstruction stability and accuracy.
VR-NeRF~\cite{VRNeRF} introduced a custom multi-camera cluster rig to densely capture real-world environments in high-resolution and high dynamic range, and used a LOD-focused rendering that can perform on a large-scale scene.
However, all these approaches only focus on the entire scene representation, neglecting the fine details of individual objects within the scene.

\textbf{Unbounded scene.} Unbounded outdoor scenes with very distant backgrounds are one of the limitations of NeRF-like methods because they require dense sampling of points along rays in space.
This would cause unbalanced details of objects near and far from the camera, resulting in blurry, low-resolution renderings.
NeRF++~\cite{Zhang2020NeRFAA} improved upon NeRF by dividing the scene into an inner unit sphere for the foreground and an outer space mapped to a more compact volume by an inverted sphere for the background.
Mip-NeRF 360~\cite{barron2022mipnerf360} is an extension of Mip-NeRF~\cite{barron2021mipnerf} that enhances performance on unconstrained scenes by using non-linear scene parameterization and a coarse-to-fine online distillation framework with a Proposal Network.
Due to its high efficiency, these unbounded properties are applied in most of the later development methods, including in the foundation of our method.

\textbf{Compositional NeRF.} Both scene synthesis and manipulation can benefit from the compositional representation of scenes based on their internal components. However, decomposing and composing NeRFs remains challenging due to the nature of their implicit representation. NeRF-like methods often learn entire scenes holistically, making it difficult to capture the semantics of specific objects or the locations of regions.
Panoptic Neural Fields~\cite{Kundu_2022_CVPR}, Nerflets~\cite{zhang2022nerfusion}, NSGs~\cite{Ost_2021_CVPR} learn multiple structured hybrid representations for urban scenes, enabling panoptic segmentation and object manipulation in NVS.
However, these category-specific methods only work on object representations that have been defined or constrained to a domain or type during or before training.
Moreover, the reconstruction fidelity is quite low and unstable due to the limited number of viewpoints available for the objects.
Other studies used object-level annotations for supervision.
ObjectNeRF~\cite{yang2021objectnerf} used a two-pathway architecture to encode scene and object radiance fields separately, thus allowing for the manipulation of objects while maintaining scene consistency.
UDC-NeRF~\cite{wang2023udcnerf} is an improvement of ObjectNeRF~\cite{yang2021objectnerf}, which integrated the decomposition and composition process into the learning model to improve the quality and consistency of the display.
However, these approaches rely heavily on view-consistent ground truth instance segmentation masks during training, which are costly to obtain in practice.
In contrast, our method does not require mask annotations and can render detailed images for user-specified ROIs, enabling NVS with high fidelity, consistency, and efficiency. 
Furthermore, built on top of Nerfacto from the Nerfstudio framework~\cite{nerfstudio}, our method can effectively handle unbounded scenes and benefit from fast training and rendering. 
Nerfacto, indeed inspired by Mip-NeRF 360~\cite{barron2022mipnerf360} and iNGP~\cite{mueller2022instant}, also integrates ideas of important improvements from works such as NeRF-W~\cite{martinbrualla2020nerfw} and Ref-NeRF~\cite{verbin2022refnerf}.

\section{Method}
\label{sect:method}
\subsection{Overview}
\label{sect:method:overview}

\begin{figure*}[!t]
\centering
\includegraphics[width=1\linewidth, height=7cm]{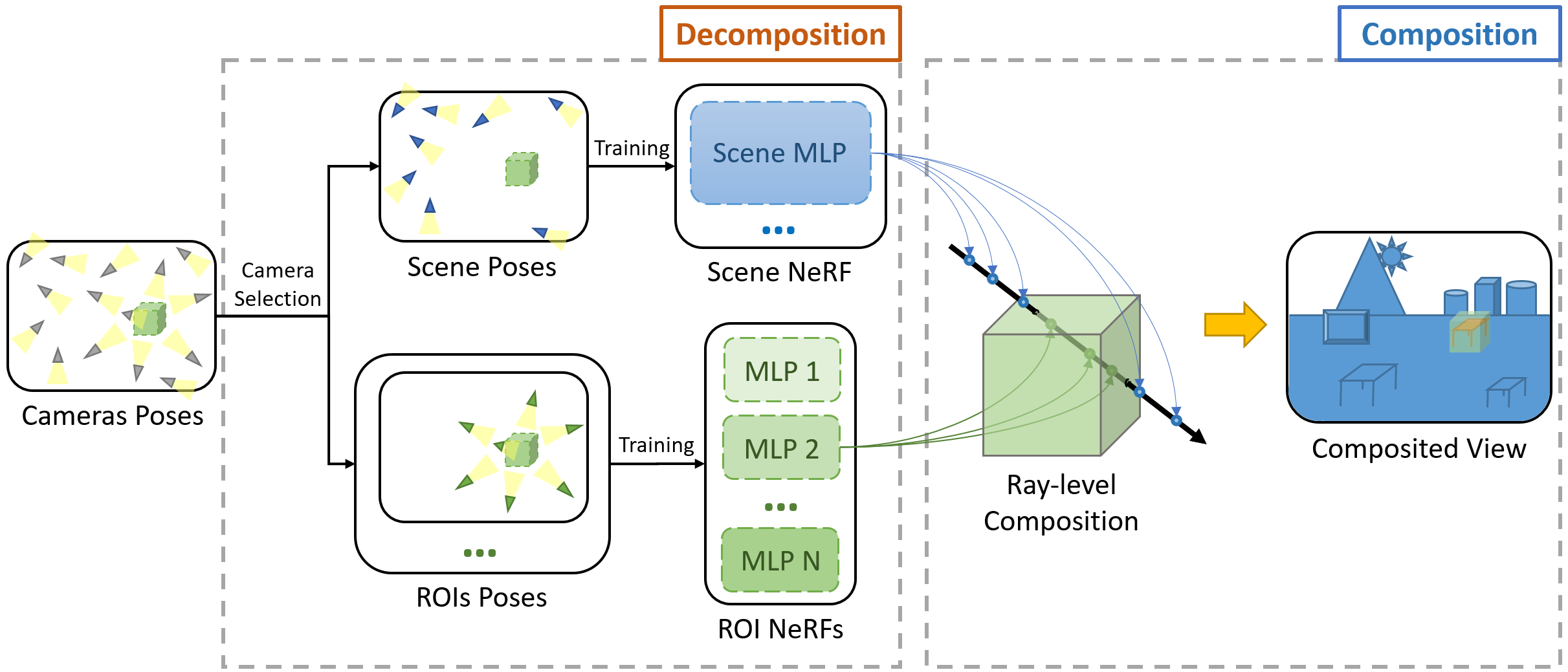}
\caption{
An overview of the proposed Region of Interest-Focused NeRFs (ROI-NeRFs) framework, consisting of two steps: scene decomposition and composition. In decomposition, the scene is divided into Scene and ROIs groups, with camera sets automatically selected for each NeRF training. 
Composition stage integrates high-detail ROI NeRFs and the global Scene NeRF to produce high-quality renderings with enhanced  detail for objects in the ROIs.
}
\label{fig:pipeline}
\end{figure*}

Our framework takes as input a set of 2D color images of a scene, along with camera poses, intrinsic parameters, and a sparse point cloud, and then produces highly realistic renderings with high LOD on objects or regions of interest.
It consists of two main stages: decomposing the scene into multiple NeRFs and compositing their inferences, as illustrated in~\cref{fig:pipeline}.

In the decomposition phase, the scene is divided into two groups: the Scene group and the ROIs group; each ROI is assigned a dedicated NeRF.
This process begins by defining ROIs using axis-aligned bounding boxes (AABBs) that closely bounds the objects of interest.
Each selected AABB isolates a specific region, enabling its NeRF to exclusively model the ROIs, ignoring the remaining scene content during training, thus ensuring high LOD within this ROI.
Next, the cameras are selected into groups for training the ROI NeRFs and the Scene NeRF. 
Cameras that are sufficiently close to an AABB and with it in their field of view are assigned to train the corresponding ROI NeRF, while the remaining cameras are used to train the Scene NeRF.
This selection process happens during the decomposition phase, see~\cref{fig:pipeline}; the automatic Object-Focused Camera Grouping technique is described in \cref{sect:method:decompo:cam_select}.
After training, we obtain one large Scene NeRF representation at a moderate LOD, along with $N$ smaller high-detail ROI NeRFs focused on the objects of interest. 

During the composition stage, the strengths of each NeRF group are leveraged for integration using our proposed Ray-level Compositional Rendering techniques, which is described in \cref{sect:method:compo}.
From an inference viewpoint, rays in space are processed to synthesize information from the NeRFs volumes they intersect.
Information within the ROIs is drawn from the high-detail ROI NeRFs, while the surrounding regions are processed by the Scene NeRF, providing broad coverage and comprehensive scene geometry, as shown in the composition part of~\cref{fig:pipeline}.
These combined rays are then used in volumetric rendering for image inference, achieving an improved display quality for specific objects of interest.

\subsection{Region of interest decomposition}
\label{sect:method:decompo}
\subsubsection{Object-Focused Camera Grouping for training}
\label{sect:method:decompo:cam_select}
In a scene, there may be one or multiple objects of interest that require close inspection or that benefit from detailed captures.
For large real-world settings, 3D reconstruction tasks demand huge datasets of thousands of images. 
Therefore, manually reviewing each image to identify ROIs for camera selection would be repetitive, subjective, time-consuming, and labor-intensive.
\begin{figure}[!t]
\centering
\includegraphics[width=0.9\columnwidth]{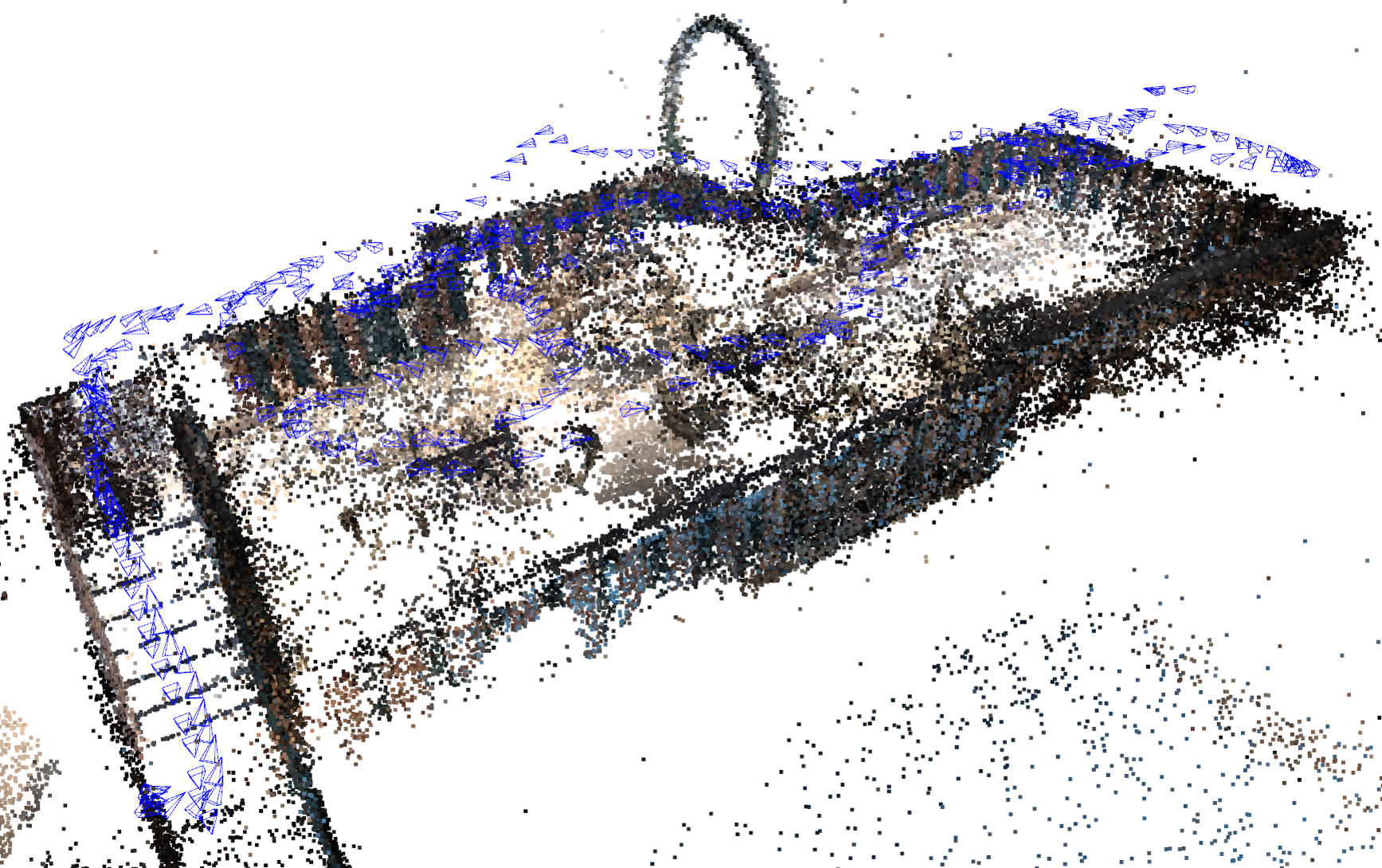}
\caption{The sparse 3D point cloud and the camera positions (in blue) estimated by SfM from the Egypt dataset.}
\label{fig:pc_campath}
\end{figure}
To streamline this process, we devised a semi-automatic selection method that leverages the visibility of the sparse 3D point cloud obtained from the SfM, such as the one from the \textit{Egypt} dataset shown in \cref{fig:pc_campath}.
This point cloud is constructed from keypoints identified in overlapping 2D images.
By analyzing these keypoints, we can select all views that capture at least one point in their 2D images. Conversely, we can also determine which keypoints are present in each camera's image.

\begin{figure}[!t]
\centering
\includegraphics[width=1\columnwidth]{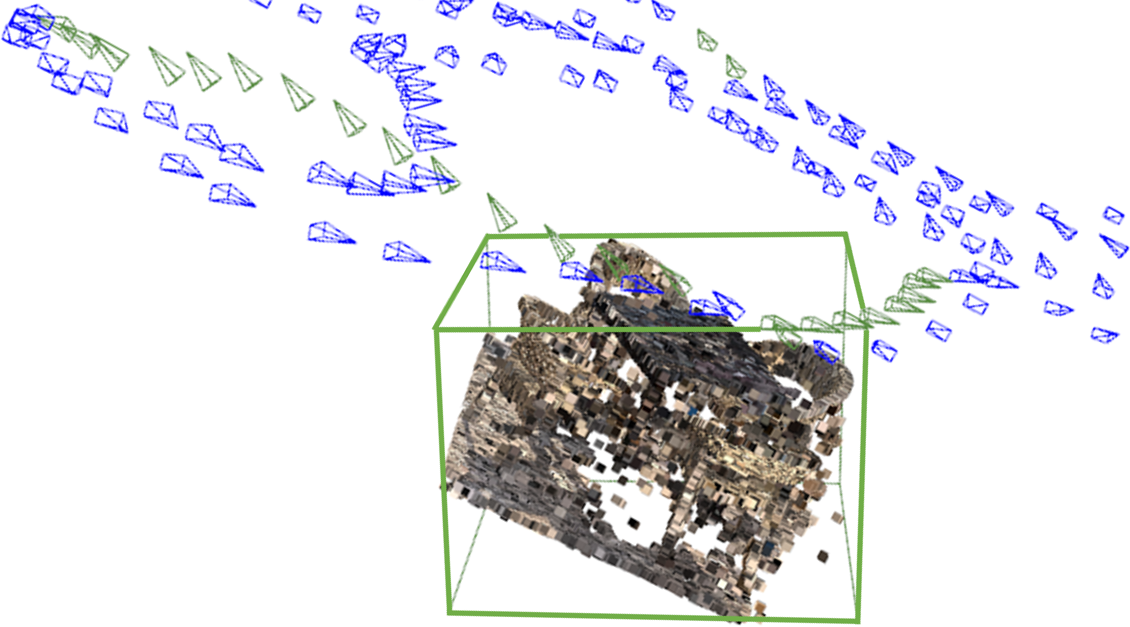}
\caption{
An example of a ROI: a table and surrounding chairs within the green AABB, manually selected from the sparse point cloud. Cameras in green focus on this ROI; they are automatically selected by the criterion of observing at least \SI{10}{\percent} of the 3D points inside the ROI. The green cameras are used to train the ROI NeRF, while the remaining blue cameras train the Scene NeRF. For clarity, only a single ROI is shown in this example.
}
\label{fig:cam_roi_selection}
\end{figure}

\textbf{Object selection from sparse point cloud.}\\
First, the objects of interest must be identified in space.
In our proof-of-concept, the AABBs surrounding the objects in the sparse point cloud are drawn by the user, as the green 3D box shown in \cref{fig:cam_roi_selection}.
Objects of interest in the point cloud are generally easy to identify due to their higher point density, which results from more cameras captures or also their inherent complexity.
As different ROIs, corresponding to multiple objects of interest, may be identified within the same scene.
We show in \cref{sect:exp:quanti_comparison} that our method naturally supports multiple ROIs and that the performance scales well when multiple ROIs are used.

\textbf{ROI-Focused Camera Grouping.}\\
From the SfM, the visibility of each point in the point cloud is given, \ie for each point, the list of corresponding 2D image points (keypoints) used for computing the 3D point is known.
For each AABB, we consider the 3D points within the bounding box and use the visibility to determine the set of cameras that captured at least $M$ keypoints in their images.
In our experiments, a typical threshold value used was $M>10\%*nbTotalPoints$, where $nbTotalPoints$ is the total number of keypoints within the AABB.
The image set is then assigned to train the corresponding ROI NeRF.
This approach is based on the observation that the number of visible keypoints correlates with the camera's distance to the ROI and the object's LOD: cameras closer to the object capture more keypoints and details, while those farther away capture fewer.
Thus, using a keypoint count filter alone is sufficient for selecting images that contribute to high-LOD training of the ROIs, without considering the camera distance.

This Camera Grouping technique ensures effective camera partitioning between the Scene NeRF and the ROI NeRFs, as illustrated in \cref{fig:cam_roi_selection}. 
It is indeed important that the Scene NeRF contains also images of the selected objects so that their geometry is effectively learned at training, to avoid issues in the subsequent ray composing process.
Therefore, a subset of the selected cameras is also integrated into the Scene NeRF training set for better scene reconstruction.

\subsubsection{NeRFs training}
\label{sect:method:decompo:training}
\begin{figure}[!t]
\centering
\includegraphics[width=0.9\columnwidth]{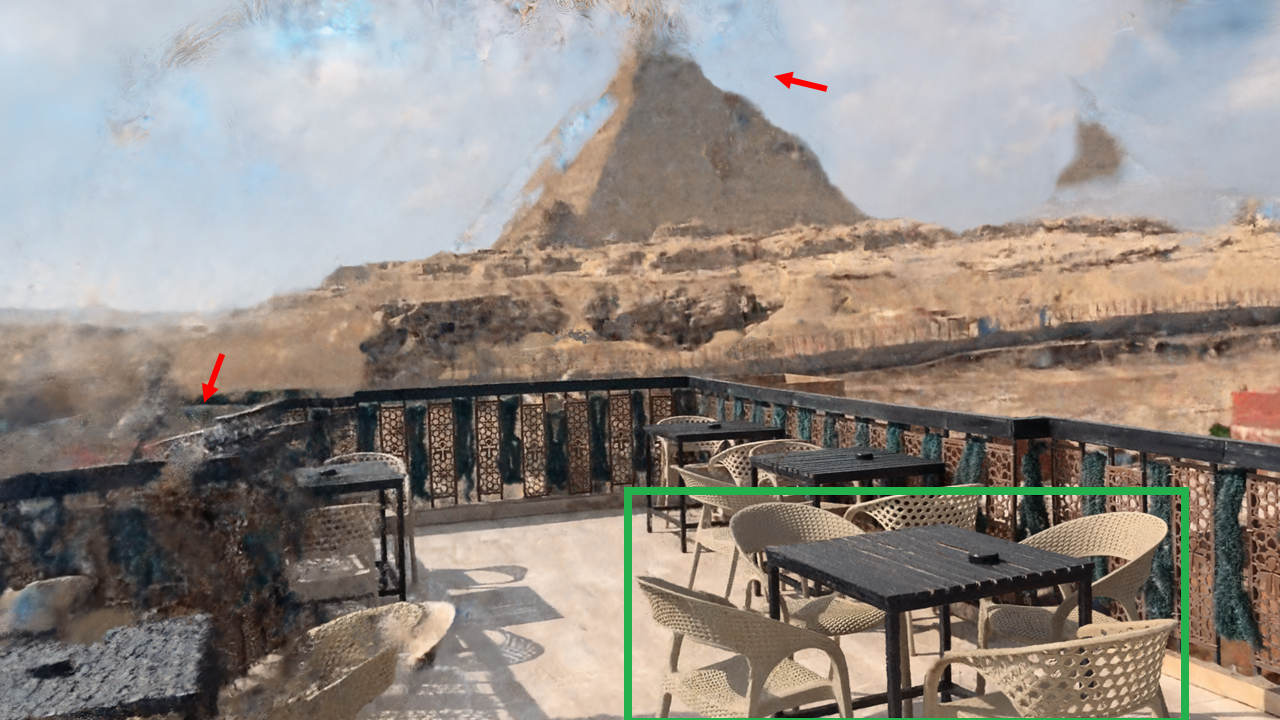}
\caption{A render of the object of interest (in the green box) and the surrounding scene from the corresponding ROI NeRF. 
The object is learned in greater detail, while the surrounding and background regions show a noticeably lower quality with floaters and missing geometry.}
\label{fig:roi_background}
\end{figure}

Our NeRF training is primarily developed on top of the default Nerfacto method of the Nerfstudio~\cite{nerfstudio} framework. 
Using the processed and grouped image sets, we train the corresponding NeRFs for Scene or ROIs groups with appropriately configured parameters, which will be further described in \cref{sect:exp:setup}.
While large-scale scene partitioning techniques like Block-NeRF~\cite{tancik2022blocknerf} or Mega-NeRF~\cite{Turki_2022_CVPR} could be applied, we employ only a single unbounded NeRF to represent the background scene in this work, avoiding the additional complexity of scene decomposition. 
The ROI NeRFs models are trained to focus on the finer details of each selected object, with an appropriate value for the finest resolution $N_{max}$ parameter, inspired from iNGP~\cite{mueller2022instant}.
This approach ensures that the ROI NeRFs achieve high detail by concentrating on a smaller volume and using high-resolution input.
After training, we obtain $N$ NeRF models for ROIs and one for the entire scene. 
As a result, the rendering quality of ROIs in these NeRFs is significantly improved compared to those in the Scene NeRF or even compared to a model trained on the entire dataset without partitioning, which we will denote \textit{Full NeRF} from now on.
However, ROI NeRFs lack information in areas outside the ROIs, thus generating images of poor quality in those areas with many artifacts and \textit{floaters}, as shown in \cref{fig:roi_background}. 
On the other hand, since it has been trained on images covering the whole scene, Scene NeRF produces better-quality images for the overall view.

\subsection{Scene composition}
\label{sect:method:compo}

In this step, we want to integrate the high LOD of the ROI NeRFs with the rest of the Scene NeRF.
A naive approach involves inferring the entire image for a given viewpoint from both the Scene NeRF and the ROI NeRFs, then using a mask of the ROI’s AABB to replace the corresponding pixels in the Scene NeRF image with those from the ROI NeRF. 
However, this method significantly increases inference time, as it requires rendering the same view multiple times, once for each NeRF, causing the rendering time to grow linearly with the number of ROIs. 
Additionally, this approach can lead to artifacts such as black or blurry spots in the ROIs, especially in views with occlusions or from camera angles quite distant from those used during training. 
These issues arise due to \textit{floaters} that block color information along the ray.
To overcome the limitations of pixel-level compositing, we developed the Ray-level Compositional Rendering technique, which processes and combines sampled points directly at the ray level.
This approach allows for more flexible compositing and improved rendering speed and image quality.

The Nerfacto model we use is inspired by Mip-NeRF 360~\cite{barron2022mipnerf360}.
It employs a small ``Proposal" MLP for point sampling along a ray, which is optimized concurrently with the main NeRF during training.
Ideally, if both the Scene NeRF and ROI NeRFs are properly learnt within the same reference system, they should generate similar point distributions along corresponding rays in the ROIs, concentrating on surface regions.
In practice, however, the point distributions between the Scene NeRF and ROI NeRFs are inconsistent as the training sets differ.

Initially, we used only the trained Scene NeRF's Proposal Network to sample points along the rays. 
Points within the object's AABBs were then fed into the ROI NeRFs for density and color information, which was then used in ray aggregation to produce the final pixel color. 
While this approach improved the display quality on the ROIs, artifacts like holes in objects sometimes persisted due to incomplete geometric information in the Scene NeRF,
as described in \cref{sect:exp:abla:RSR} . 
To mitigate these issues, we refined our Ray-level Compositional Rendering with the following key strategies.

\begin{figure}[!t]
\centering
\includegraphics[width=1\columnwidth]{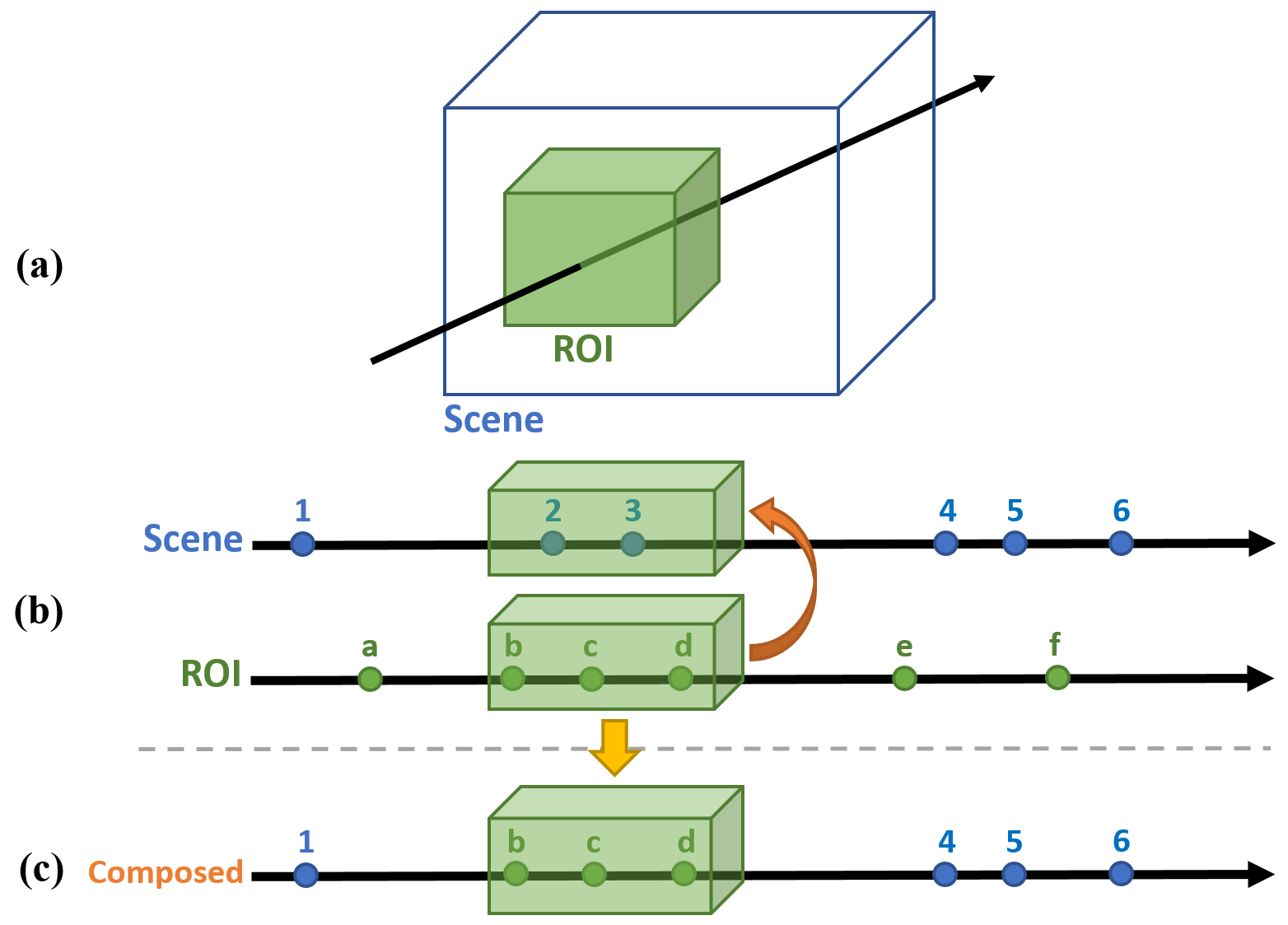}
\caption{Ray-level Composition. 
\textbf{(a)} For a given ray passing through the ROI, \textbf{(b)} the sampled points from the ROI and Scene NeRF differ.
Since both rays are within the same normalized space, the bounding intervals on the rays inside the AABB are identical, represented by the green boxes.
\textbf{(c)} We replace all Scene points within the AABB with points sampled from the ROI NeRF to obtain the composed ray. 
}
\label{fig:ray_compo}
\end{figure}

\textbf{Scene samples replacement with ROI samples.}\\
Instead of using only the Scene sampler for all calculations, the ROI samplers are used concurrently.
Points sampled from the Scene NeRF within the AABBs are replaced with those from the ROI NeRFs, as shown in \cref{fig:ray_compo}, ensuring a more accurate representation of the object.
This approach minimizes hole artifacts and improves quality, but introduces a new issue as the numbers of points on the composed rays are uneven. 
To address this, we implemented a technique called Uniform Ray Composition, detailed in the Supplementary Material.

\begin{figure}[!t]
\centering
\includegraphics[width=1\columnwidth, height=6cm]{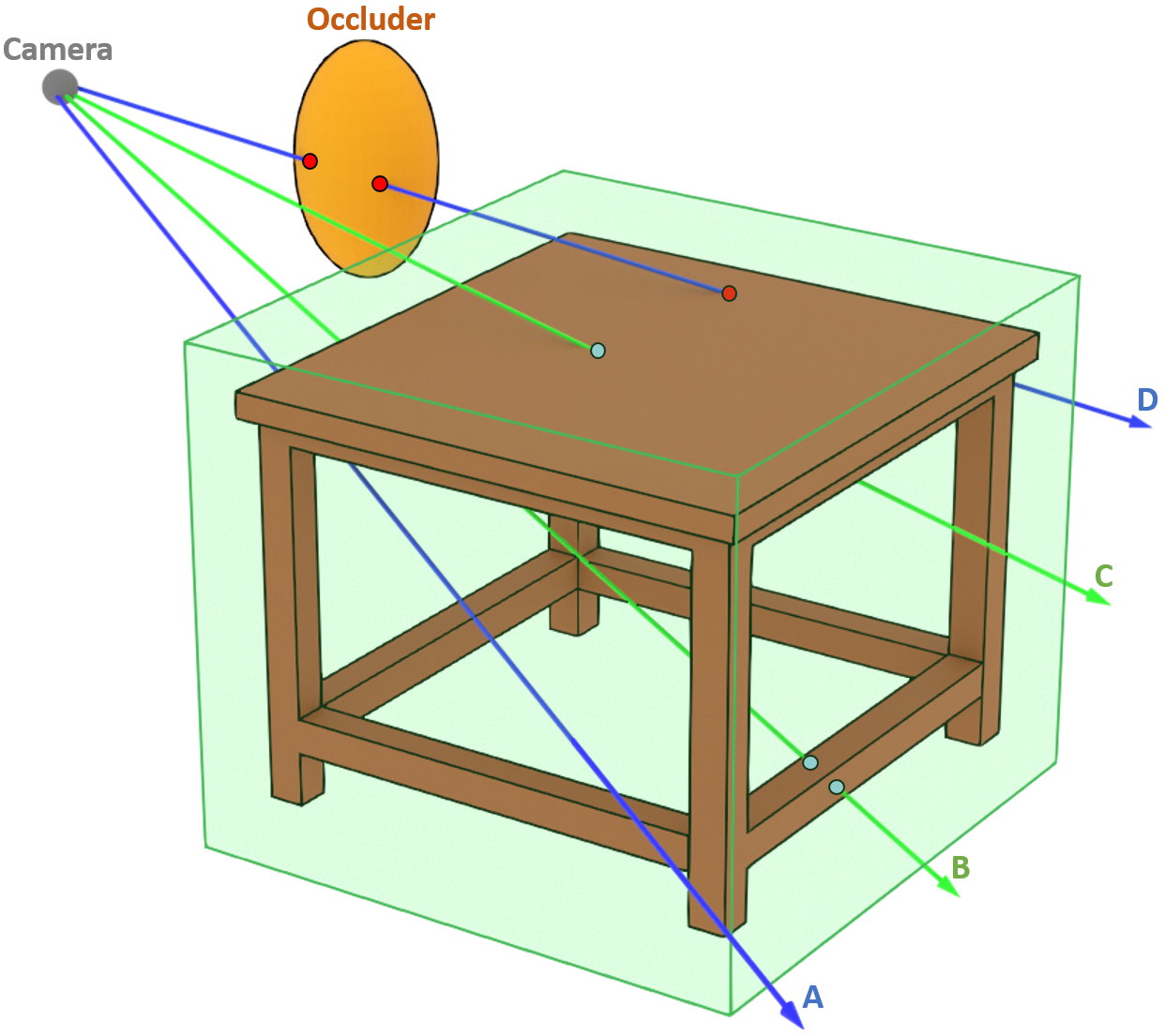}
\caption{A visualization of possible ray types passing through the ROI, represented here as a table. Blue ray \textcolor{blue}{A} traverses empty spaces within the box without intersecting any part of the table. 
Blue ray \textcolor{blue}{D} interacts with the tabletop but is blocked by another object between the table and the camera. 
Green rays \textcolor{ForestGreen}{B} and \textcolor{ForestGreen}{C} directly intersect the tabletop and the thin horizontal bar.
In summary, blue rays are redundant and should be filtered out to optimize the compositing process.
}
\label{fig:roi_rays}
\end{figure}

\textbf{ROI Depth-based Rays Filtering.}\\
Not all ROI samples within an AABB are useful for improving the LOD. 
For example, points in empty spaces contribute little to improving compositional quality. 
Although a properly learned Proposal Network typically minimizes the presence of empty sampled points, some rays can intersect the box without hitting any significant geometry.
Additionally, there are instances where an ROI's box is within the camera's view frustum but is occluded by other objects. 
Consider the rays passing through the AABB in \cref{fig:roi_rays}. 
Ray A is an example of an empty ray that travels through space without intersecting the surface of any object in the ROI.
Ray D, although it intersects the table surface, is blocked by another object before reaching the ROI being processed.
Processing rays in these empty or occluded regions is inefficient and increases computation time.
Therefore, the proposed method uses the geometric information learned from training to filter out these non-contributing rays.
Indeed, the depth map for rays passing through the AABBs is computed from the corresponding trained ROI NeRF, as it provides a more accurate representation of the geometry and appearance of the content within the AABB compared to the Scene NeRF.
Each ray passing through the ROI is then assigned an associated depth.
This is achieved through a depth calculation feature in Nerfstudio~\cite{nerfstudio}, where densities along the rays are aggregated, and the depth is set at the distance where the cumulative ray weight reaches $0.5$.
If a ray's depth falls outside the ROI (either in front or behind), it is filtered out, and the ray is instead rendered by the Scene NeRF or another NeRF object that occludes the current ROI.
Furthermore, the densities and feature vectors of the ray samples calculated during this step are leveraged to efficiently compute the color for the ray composition process later on, avoiding redundant queries to the MLP.
By applying this Depth-based Rays Filtering, we have significantly reduced the number of rays requiring composition, slightly improved rendering quality, minimized artifacts in empty regions within the AABB and ROI boundaries, as well as simplified the implementation of multiple ROIs composition simultaneously.

\textbf{Handling distant objects of interest.}\\
ROIs beyond a certain distance $d_{max}$ from the camera are excluded from composition. The distance $d_{max}$ is defined as the distance from a corresponding ROI’s center to the farthest camera in the training set. These distant ROIs are inferred from the Scene NeRF at medium detail, improving inference speed and reducing unnecessary processing for negligible improvements.
When rendering videos, objects near $d_{max}$ may show a popping effect due to sudden changes in detail. Future improvements could include blending Scene and ROI inferences across a range of distances around $d_{max}$ or setting $d_{max}$ so transitions are imperceptible at the rendered resolution.

\section{Experiments}
\label{sect:exp}
\subsection{Experimental setup}
\label{sect:exp:setup}

In this Section, we discuss the implementation details of our method and present the results providing quantitative and qualitative evaluations of our method on two real-world datasets.

The \textbf{Egypt} dataset is a real-world dataset available in Nerfstudio~\cite{nerfstudio}, comprises $302$ full-HD images captured on a café rooftop overlooking a pyramid in Egypt.
The dataset primarily focuses on tables, chairs, and decorative elements, with the pyramid, city buildings, and blue sky as the background.

The \textbf{Hôtel de la Marine} dataset was captured by our lab at the cultural heritage site Hôtel de la Marine in Paris\footnote{\url{https://www.hotel-de-la-marine.paris/}}.
This dataset will be publicly released with the publication of the paper.
It consists of $916$ images at $8192\times5464$ resolution taken with a professional Canon camera using a \SI{24}{\milli\metre} wide-angle lens, and $222$ close-up images at $3024\times4032$ resolution captured with an iPhone 15 Pro Max. 
The dataset walks through an entire hotel dining room, including close-ups of objects on a small table, such as a fruit plate, lamp, flower basket, \etc.
This dataset is a good example of the kind of applications our method targets.
It offers the user the possibility to visualize many different objects of interest with historical value.
To reduce storage and computational costs, image resolutions were downscaled by four $4$ to $2048\times1366$ and $756\times1008$, respectively.

\begin{figure*}[!t]
    \centering
    \subfloat[]{
        \includegraphics[width=0.5\textwidth]{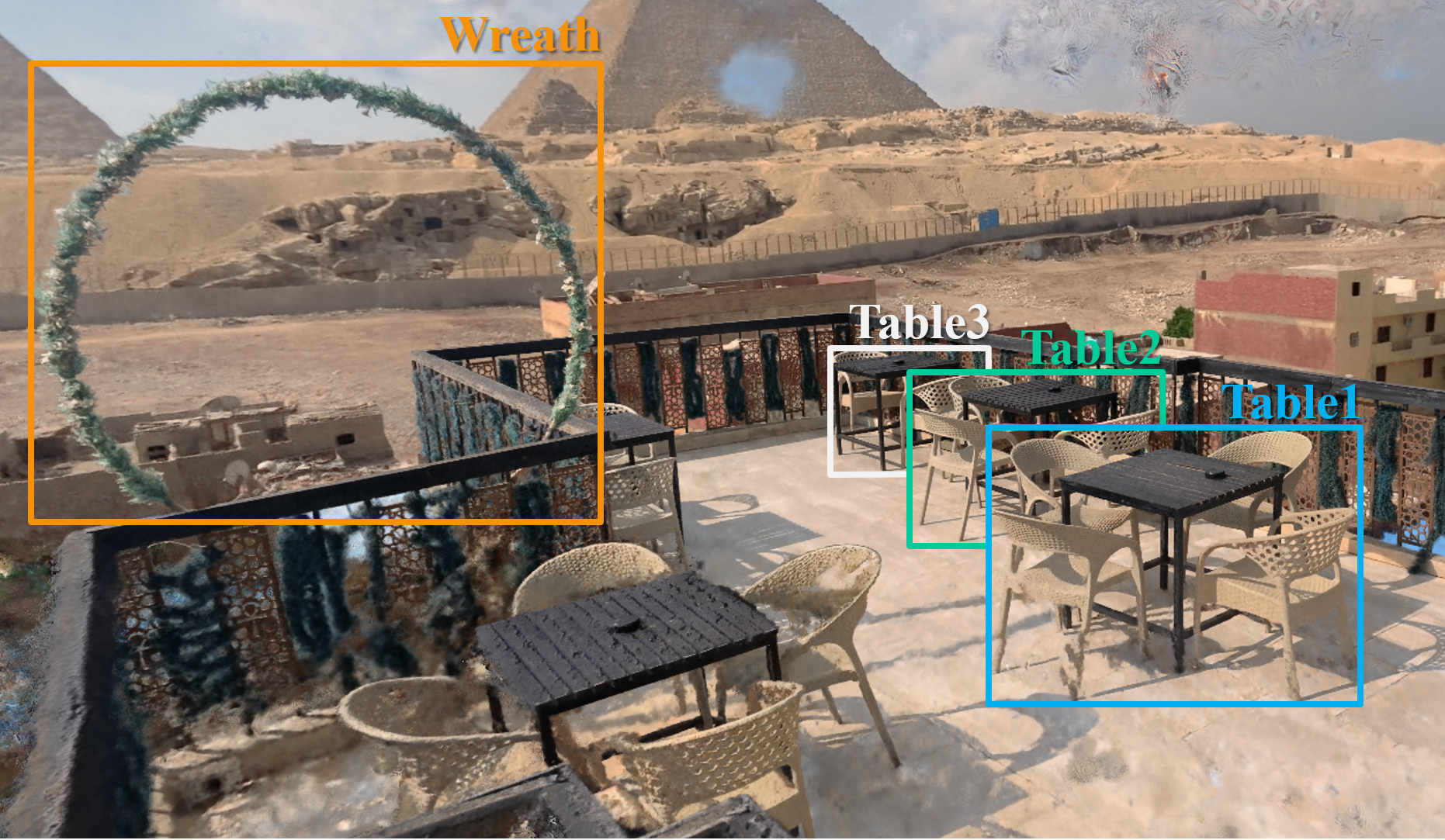}
        \label{fig:rois_egypt}
    }
    \hfil
    \subfloat[]{
        \includegraphics[width=0.452\textwidth]{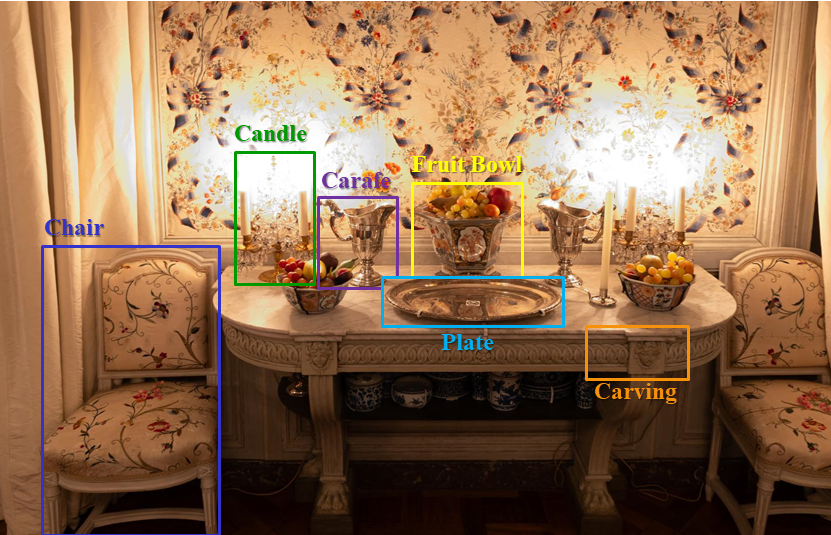}
        \label{fig:rois_hotel}
        }
    \caption{
    An illustration of selected objects of interest in the \textbf{(a)} Egypt and \textbf{(b)} Hôtel de la Marine dataset. Objects are marked with colored boxes along with their short names which we will use to refer to them from now on.
    }
    \label{fig:object_illust}
\end{figure*}

For both datasets, we used Agisoft Metashape\footnote{\url{https://www.agisoft.com/}} to calibrate and retrieve the camera poses, as well as the sparse point cloud.
In the \textit{Egypt} dataset, we selected $4$ objects of interest: $3$ sets of table and surrounding chairs and a decorative tree wreath, as shown in \cref{fig:rois_egypt}.
In the \textit{Hôtel de la Marine} dataset, we selected $6$ objects of interest located in the table area with available close-up images, as shown in \cref{fig:rois_hotel}.
The ROI AABBs were then used in the Camera Grouping process described earlier to obtain the corresponding training sets.
For each ROI, we select the cameras in which at least \SI{10}{\percent} of the 3D points inside the ROI are visible.
Additionally, \SI{50}{\percent} of the ROI's camera training sets were also integrated into the Scene NeRF training set, to retain the information of the ROIs when learning the Scene NeRF.
This Object-Focused Camera Selection process also simulates a scenario where the Scene NeRF does not have access to detailed images of objects. 
This setup can also be used to evaluate potential display quality improvements when additional detailed images are captured in the future.
For each ROI NeRF, $\sim$\SI{15}{\percent} of the grouped cameras are randomly selected and evenly distributed around the object for the test set, ensuring evaluation images always see the corresponding ROI, regardless of its size.

\subsubsection*{Implementation details}
\label{sect:exp:implementation}

Our implementation is primarily based on the Nerfstudio\footnote{\url{https://github.com/nerfstudio-project/nerfstudio/}} framework.
All calculations are performed using a single \SI{80}{\giga\byte} A100 GPU on a DGX station.

For training, we used the default Nerfacto model for the ROI NeRFs.
Besides, we employed larger versions of the Nerfacto model, Nerfacto-big and Nerfacto-huge, with larger parameters such as hash table size $T$, finest grid resolution $N_{max}$, and number of iterations.
These models are better suited for encoding the entire surrounding scene.
To demonstrate good adaptability and flexibility of our approach, comparisons are made with the Scene NeRF, trained using both ``big'' and ``huge'' models.
Additionally, we trained a Full NeRF, a version of Scene NeRF, on the entire image set without any camera partitioning for a more thorough comparison. 

For ROI NeRFs training, in addition to the default parameters, we estimated the maximum possible resolution ($N_{max}$) for the highest grid level that the model can encode. 
This estimation is based on factors such as the image resolution, camera positions, and ROI's AABBs locations. 
We calculated $N_{max}$ by considering the camera position closest to the ROI box, ensuring that the smallest unit of the image, the pixel, corresponds to the smallest unit of the model's finest 3D grid, the voxel.
This approach enables the model to capture the highest LOD without setting an excessively high $N_{max}$.
For the Scene NeRF, determining an appropriate $N_{max}$ for the entire scene is challenging, so we retained the default $N_{max}$ of $4096$ for Nerfacto-big and $8192$ for Nerfacto-huge, as the required LOD is medium.
Similarly, the model size, determined by the hash table size $T$, was kept at the default values -- $2^{19}$ for ROIs and $2^{21}$ for the Scene.

\subsection{Quantitative evaluation} 
\label{sect:exp:quanti_comparison}
We utilized standard metrics such as PSNR, SSIM, and LPIPS for consistent and meaningful comparisons.
The baselines used in all subsequent evaluations include the Scene NeRF and Full NeRF, each of them available in two versions according to the training settings: ``big'' (Scene NeRF-big ($1$) and Full NeRF-big ($2$)) and ``huge'' (Scene NeRF-huge ($3$) and Full NeRF-huge ($4$)).
These baselines were trained with the default $100$k iterations to ensure optimal learning of scene details. 
Heuristically, training the entire scene with a larger baseline model or even with more training images to improve local detail quality can be both costly and inconsistent.
In this section, we evaluate our method in two versions: single- and multiple- composition. 
The single-composition approach focuses on a single corresponding ROI at a time within its selected test set. 
The multiple-composition approach combines all valid ROIs in the inference view across the same test images used by the single-composition and baseline methods, ensuring comparability and fairness. 
The metric scores reported in the tables are computed over the entire image, not just in the object of interest regions, providing a more general and holistic evaluation of the methods.

\begin{table*}[ht!] 
    \centering
    \caption{
    Overall evaluations of  our methods compared to baselines (1) to (4) across all ROIs in the two datasets. 
    The proposed composition methods, performed on the corresponding ROI or all valid ROIs, yield results averaged across all ROIs in each dataset. 
    Our composition methods show consistent performance improvement over the baselines. Especially the multiple-composition method consistently outperforms and improves the rendering quality.
    }
    \label{tab:quanti_mean}
    \resizebox{1\textwidth}{!}{
        \begin{tabular}{l|*{6}{c}|*{6}{c}}
        Dataset & \multicolumn{6}{c|}{Egypt} & \multicolumn{6}{c}{Hôtel de la Marine} \\ 
        Methods$\vert$Metrics
        & {PSNR\textsuperscript{$\uparrow$}} & {SSIM\textsuperscript{$\uparrow$}} & {LPIPS\textsuperscript{$\downarrow$}} & {Train} & {Render} & {Mem}
        & {PSNR\textsuperscript{$\uparrow$}} & {SSIM\textsuperscript{$\uparrow$}} & {LPIPS\textsuperscript{$\downarrow$}} & {Train} & {Render} & {Mem}\\
        \hline
        \hline
        ROI NeRF (0)            & 20.23 & 0.697 & 0.316 & 19m & 2.17s & 177MB       
                                                                                & 20.94 & 0.686 & 0.373 & 24m & 0.95s & 173MB \\
        \hline
        \hline
        Scene NeRF-big (1)      & 21.69 & 0.717 & 0.293 & 2h13m & 4.72s & 546MB         
                                                                                & 20.37 & 0.670 & 0.418 & 3h26m & 2.75s & 547MB \\
        Ours single: (1) + (0)                 & \cellcolor{yellow!50}{21.86} & \cellcolor{yellow!50}{0.724} & \cellcolor{yellow!50}{0.282} & - & 5.47s & -      
                                                                                & \cellcolor{yellow!50}{20.45} & \cellcolor{yellow!50}{0.671} & \cellcolor{yellow!50}{0.412} & - & 3.26s & - \\
        Ours multiple: (1) + (0)s       & \cellcolor{orange!50}{22.06} & \cellcolor{orange!50}{0.731} & \cellcolor{orange!50}{0.274} & - & 6.31s & -         
                                                                                & \cellcolor{orange!50}{20.59} & \cellcolor{orange!50}{0.672} & \cellcolor{orange!50}{0.402} & - & 4.80s & - \\
        \hline
        Full NeRF-big (2)       & 22.58 & 0.744 & 0.262 & 2h14m & 4.71s & 546MB        
                                                                                & 20.26 & 0.664 & 0.429 & 3h46m & 2.75s & 547MB \\
        Ours single: (2) + (0)                 & \cellcolor{yellow!50}{22.65} & \cellcolor{yellow!50}{0.747} & \cellcolor{yellow!50}{0.256} & - & 5.46s & -      
                                                                                & \cellcolor{yellow!50}{20.36} & \cellcolor{yellow!50}{0.665} & \cellcolor{yellow!50}{0.421} & - & 3.26s & - \\
        Ours multiple: (2) + (0)s       & \cellcolor{orange!50}{22.80} & \cellcolor{orange!50}{0.752} & \cellcolor{orange!50}{0.251} & - & 6.30s & -         
                                                                                & \cellcolor{orange!50}{20.53} & \cellcolor{orange!50}{0.668} & \cellcolor{orange!50}{0.406} & - & 4.80s & - \\
        \hline
        \hline
        Scene NeRF-huge (3)     & 22.02 & 0.736 & 0.256 & 2h49m & 4.44s & 585MB        
                                                                                & 20.47 & 0.682 & 0.369 & 4h5m & 2.59s & 586MB \\
        Ours single: (3) + (0)                & \cellcolor{yellow!50}{22.17} & \cellcolor{yellow!50}{0.742} & \cellcolor{yellow!50}{0.251} & - & 5.12s & -      
                                                                                & \cellcolor{yellow!50}{20.54} & \cellcolor{yellow!50}{0.683} & \cellcolor{yellow!50}{0.367} & - & 3.03s & - \\
        Ours multiple: (3) + (0)s       & \cellcolor{orange!50}{22.32} & \cellcolor{orange!50}{0.747} & \cellcolor{orange!50}{0.246} & - & 5.92s & -         
                                                                                & \cellcolor{orange!50}{20.68} & \cellcolor{orange!50}{0.686} & \cellcolor{orange!50}{0.366} & - & 4.47s & - \\
        \hline
        Full NeRF-huge (4)      & 22.50 & 0.756 & 0.235 & 2h50m & 4.43s & 585MB         
                                                                                & 20.37 & 0.668 & 0.399 & 4h27m & 2.59s & 586MB \\
        Ours single: (4) + (0)                 & \cellcolor{yellow!50}{22.57} & \cellcolor{yellow!50}{0.757} & \cellcolor{yellow!50}{0.234} & - & 5.10s & -      
                                                                                & \cellcolor{yellow!50}{20.46} & \cellcolor{yellow!50}{0.669} & \cellcolor{yellow!50}{0.395} & - & 3.02s & - \\
        Ours multiple: (4) + (0)s       & \cellcolor{orange!50}{22.71} & \cellcolor{orange!50}{0.761} & \cellcolor{orange!50}{0.232} & - & 5.91s & -         
                                                                                & \cellcolor{orange!50}{20.64} & \cellcolor{orange!50}{0.671} & \cellcolor{orange!50}{0.389} & - & 4.47s & - \\

        \end{tabular}
    }
\end{table*}

\cref{tab:quanti_mean} presents the quantitative evaluations of our method, comparing each baseline version $(1)$ -- $(4)$ with our compositional approach, which combines the ROI NeRF with the baseline models. It provides the average model evaluations across all ROIs, including both our single- and multiple- composition approaches.
The ROI NeRFs were trained for $30$k iterations and produced higher-detail on ROI inference. 
However, they lacked context about the surrounding scene, leading to theoretically worse scores compared to baseline methods on test sets with sufficiently distant cameras, where the ROI’s occupied area in the image is not too big.

Regarding the performance of our proposed method, as seen in \cref{tab:quanti_mean}, both the single- and multiple- composition versions consistently improve object rendering quality across all four baseline models, even for the Full NeRF model trained on the entire dataset, which includes the images used for training the ROI NeRF alone.
For image inference, our method's overall performance is largely influenced by the quality of the background Scene NeRF. 
Improvements over the baseline depend on whether the ROIs are within view, sufficiently close, or occupy a significant area in the rendered image. 
Some test images show up to a \SI{1}{\decibel} enhancements in PSNR (see Table \RomanNumeralCaps{4} in the Supplementary Material). 
However, if an ROI is within the frustum but too far from the viewpoint, we exclude its composition to minimize rendering time, as the quality improvement is negligible. 
In such cases with no valid ROIs visible, no improvement is observed compared to the baseline.

Finally, the training and inference times vary due to differences in the number and resolution of images in the two datasets.
As our downstream approach uses already trained models, the training time for our methods can be considered as the combined training times of the component NeRFs.
According to the image inference time, our single-composition method introduces a slight increase in rendering time compared to the baseline due to the integration of additional processing techniques.
However, by optimizing the samples in 3D space to prevent duplication and avoiding the need to infer all NeRF ROIs for every pixel, the extra processing time does not scale linearly with the number of ROIs composed in the scene. 
Indeed, while the multiple-composition method shows an increase in rendering time compared to the single-composition method, this increase is not proportional to the number of combined objects, which can be up to six in the \textit{Hôtel de la Marine} dataset.
We invite the reader to refer to the supplementary material for more details.

\subsection{Qualitative evaluation} 
\label{sect:exp:quali_comparison}

\begin{figure*}[ht!]
	\centering
	\includegraphics[width=1\linewidth]{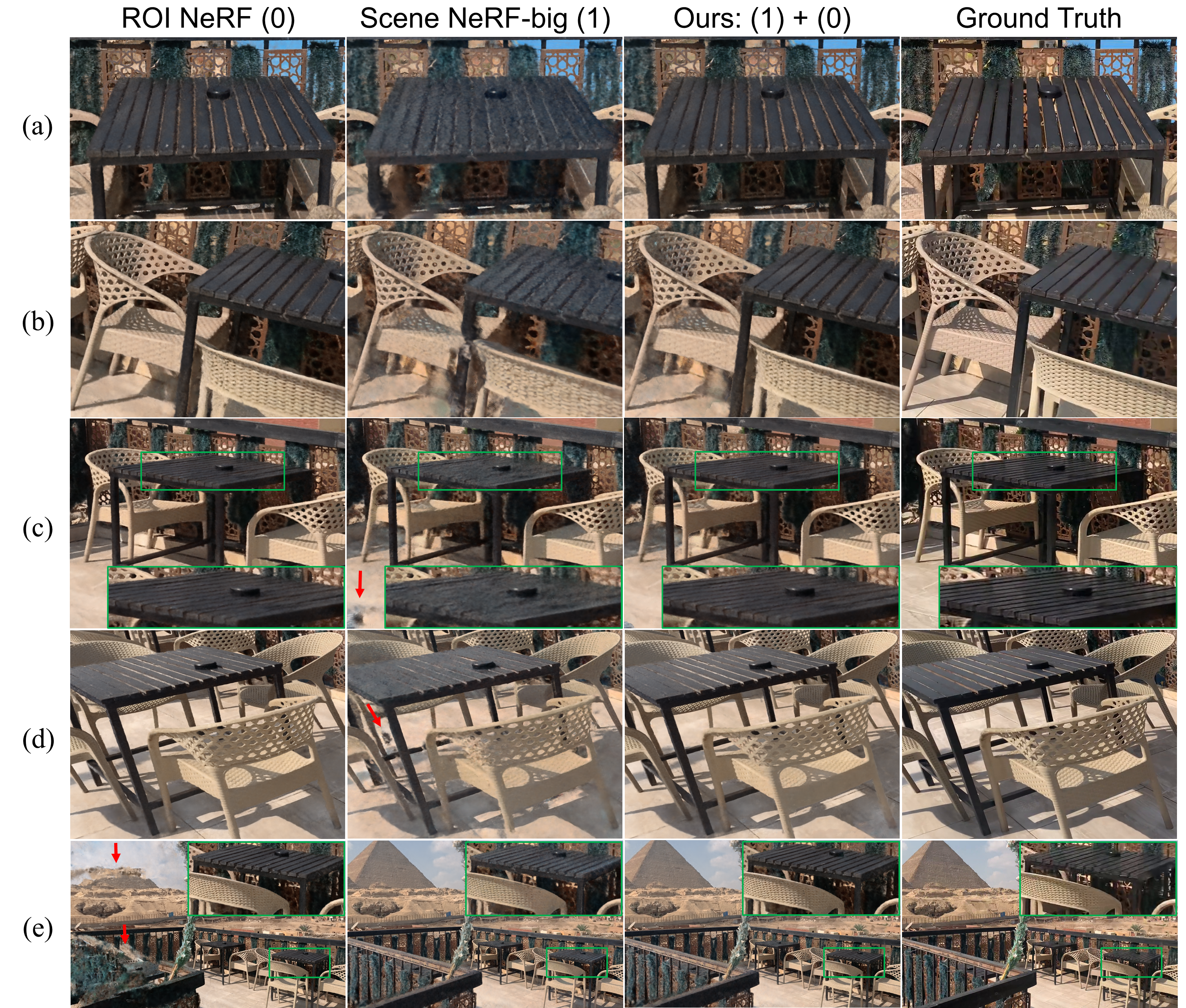}
	\caption[]{A visual comparison of the proposed method and the baseline Scene NeRF-big (1) on five ground-truth views (rows) from the retained test images of the \textit{Egypt} dataset. Views (a), (b), and (e) focus on $\mathit{Table2}$, view (c) towards $\mathit{Table3}$, and view (d) centers on $\mathit{Table1}$.
 Views (a)-(d) range from medium to close-up perspectives, while view (e) provides an overview. Our compositing method shows clear enhancements in detail within the ROIs and also reduces artifacts and floaters.
 Subtle quality differences are highlighted with green insets, while blurred areas and floaters are indicated by red arrows.}
	\label{fig:egypt_comparisons}
\end{figure*}

\begin{figure*}[ht!]
	\centering
	\includegraphics[width=0.94\linewidth]{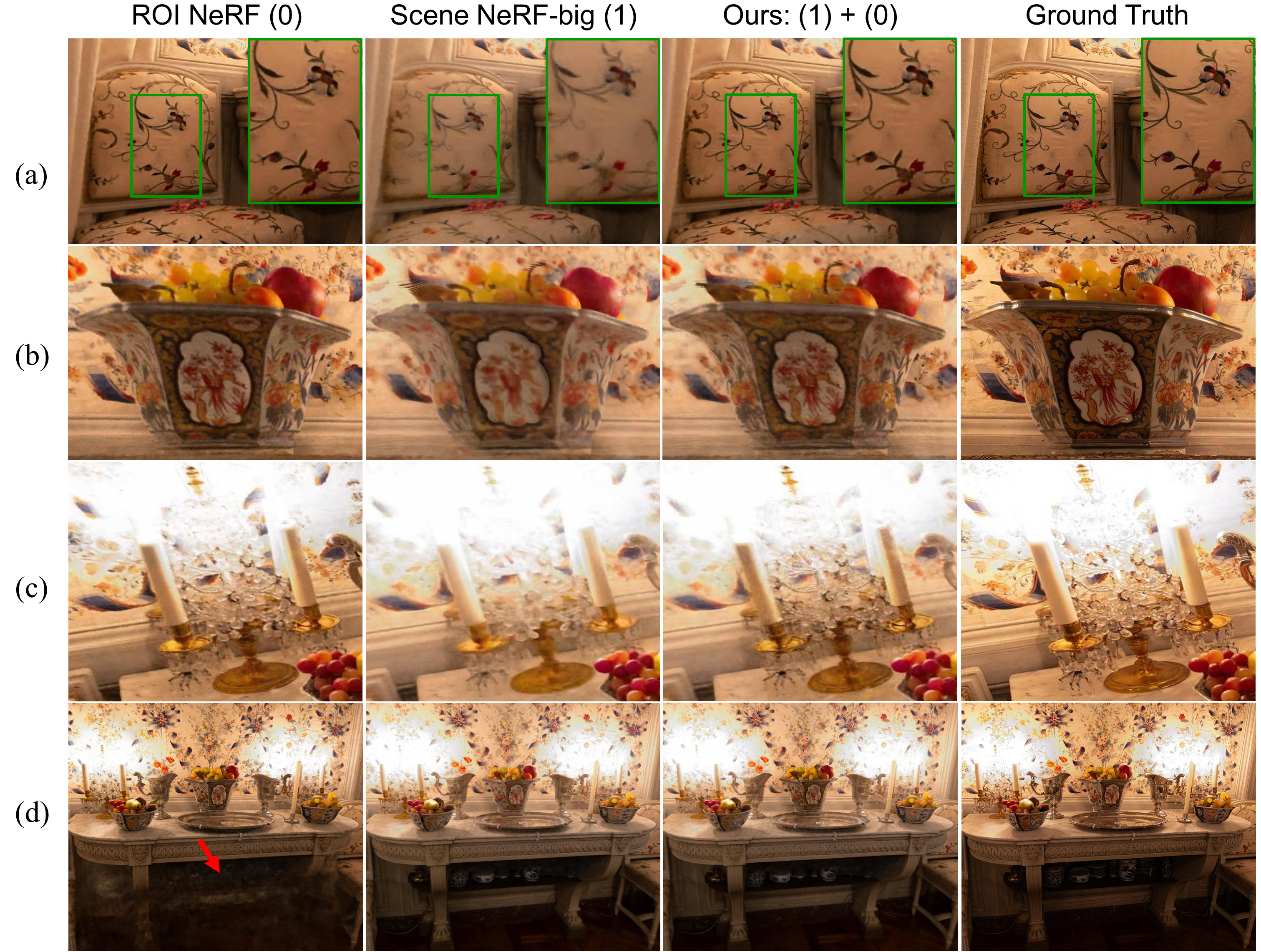}
	\caption[]{Visual comparison of our proposed method and the baseline Scene NeRF-big (1) on the \textit{Hôtel de la Marine} dataset. 
 View (a) focuses on the $\mathit{Chair}$, views (b) and (d) on the $\mathit{FruitBowl}$, and views (c) on the $\mathit{Candle}$. Views (a)–(c) show close-up perspectives, while views (d) provide overviews. Subtle quality differences are highlighted with green insets, while blurred areas and floaters are indicated by red arrows.
 }
	\label{fig:hotel_comparisons}
\end{figure*}

We also provide visual comparisons between our method and the baseline on the left-out test images in \cref{fig:egypt_comparisons} and \cref{fig:hotel_comparisons} for the \textit{Hôtel de la Marine} dataset. 
For the \textit{Egypt} dataset \ref{fig:egypt_comparisons}, the ROI NeRFs learn fine details effectively in the regions of interest but produce blurred inferences with artifacts in other areas of the scene due to lack of information. 
This is evident in the general view in the (e) row of the ROI NeRF (0), particularly in the background Pyramid and the balcony area.
In contrast, Scene NeRFs offer a comprehensive overall geometric and appearance definition for the entire scene, but local detail remains moderate to low. 
In the close-up views in the (a) and (b) rows of the Scene NeRF-big (1), the tabletop and chair mesh appear blurry and also show less geometric stability compared to the ROI NeRF (0).
Our method, illustrated in the third column, combines the strengths of both models, achieving high level of detail for objects such as tabletop and chair mesh surfaces within the ROI. 
Additionally, artifacts and \textit{floaters} in the ROIs of the Scene NeRF-big (1), indicated by red arrows in the (c) and (d) rows, are effectively handled by our compositional approach.
Similar observations are made in the \textit{Hôtel} dataset \ref{fig:hotel_comparisons}. 
While the ROI NeRFs (0) captures fine details within the ROI's AABBs, it lacks geometric accuracy in other regions, such as the region under the table as shown in row (d) of \cref{fig:hotel_comparisons}. 
Conversely, the Scene NeRF (1) provides better generalization but produces blurry details in the ROI. 
Our composition approach effectively combines these strengths, achieving high level of detail for objects like the $\mathit{Chair}$ textures or $\mathit{FruitBowl}$'s patterns while preserving overall scene generalization.

\subsection{Ablations} 
\label{sect:exp:abla}
In this section, we show the impact of each component of our method: ROI Samples Replacement (RSR) and Depth-based Rays Filtering (DRF). 
The quantitative impacts of each contribution are summarized in \cref{tab:ablation2}.
\begin{table}[h] 
    \centering
    \caption{PSNR scores for Ablations. This study is conducted on two ``big'' baseline methods and their compositions with single or multiple ROIs across all ROIs in both datasets. The experiments include: (a) No ROI Samples Replacement (RSR) and no Depth-based Rays Filtering (DRF), using only Scene samples for composition, (b) No RSR, with DRF only, (c) RSR without DRF, and (d) our complete model. 
    }
    \label{tab:ablation2}
    \resizebox{0.5\textwidth}{!}{
        \begin{tabular}{l|*{1}{c}|*{1}{c}|*{1}{c}|*{1}{c}}
        Dataset & \multicolumn{2}{c|}{Egypt} & \multicolumn{2}{c}{Hôtel de la Marine} \\ 
        Methods$\vert$Mode
        & {Single} & {Multiple} & {Single} & {Multiple} \\

        \hline
        Scene NeRF-big (1)      & \multicolumn{2}{c|}{21.69} & \multicolumn{2}{c}{20.37}\\
        \hline
        No RSR, No DRF (a)      & 21.73 & 21.76 & 20.43 & 20.57 \\
        No RSR (b)              & 21.71 & 21.73 & 20.43 & 20.54 \\
        No DRF (c)              & \cellcolor{orange!50}{21.89} & /     & \cellcolor{yellow!50}{20.44} & / \\
        Full (d)                & \cellcolor{yellow!50}{21.86} & \cellcolor{orange!50}{22.06} & \cellcolor{orange!50}{20.45} & \cellcolor{orange!50}{20.59} \\

        \hline
        Full NeRF-big (2)       & \multicolumn{2}{c|}{22.58} & \multicolumn{2}{c}{20.26}\\
        \hline
        No RSR, No DRF (a)      & 22.46 & 22.39 & 20.36 & 20.51 \\
        No RSR (b)              & 22.46 & 22.39 & 20.35 & 20.48 \\
        No DRF (c)              & \cellcolor{yellow!50}{22.64} & /     & \cellcolor{yellow!50}{20.36} & / \\
        Full (d)                & \cellcolor{orange!50}{22.65} & \cellcolor{orange!50}{22.80} & \cellcolor{orange!50}{20.37} & \cellcolor{orange!50}{20.53} \\

        \end{tabular}
    }
\end{table}

\subsubsection{ROI Samples Replacement}
\label{sect:exp:abla:RSR}

\begin{figure}[ht]
	\centering
	\includegraphics[width=1\linewidth]{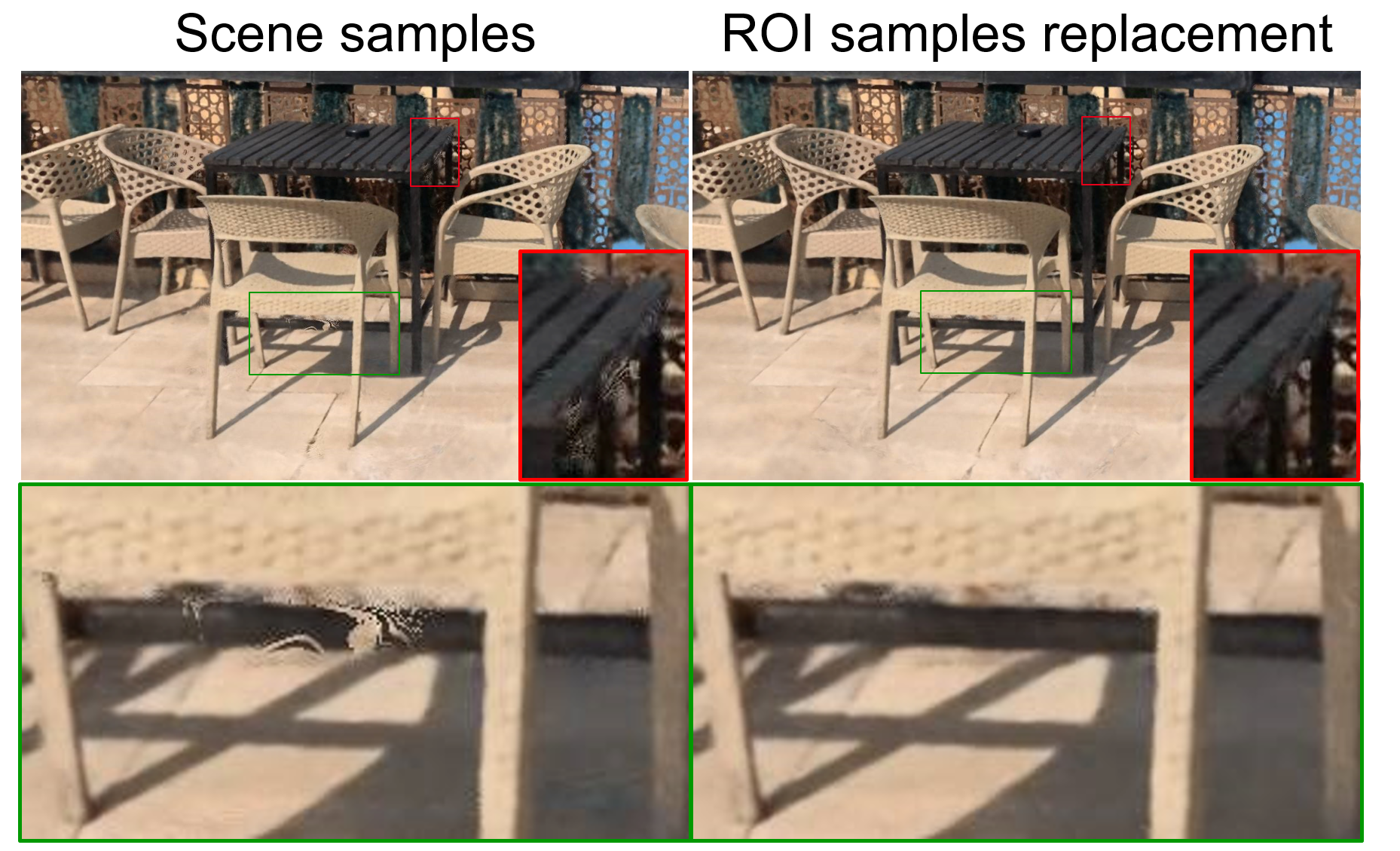}
	\caption[]{
Comparison of rendering computed with points sampled from Scene NeRF (left column) and rendering computed with points inside the AABB replaced by samples from ROI.
On the left, hole artifacts appear on the thin bottom bar and along the outer edge of the table. 
ROI sample replacement effectively ``fills'' these holes, significantly improving the display quality within the ROI. 
The artifacts and the improvements are highlighted with insets for optimal visualization.
}
	\label{fig:abla_RSR}
\end{figure}

We evaluate the importance of the inside AABB ROI sample replacement strategy. 
When training large scenes using only Scene NeRF, the model often struggles to learn local fine details, potentially leading to missing information in thin components or object boundaries, resulting in holes or noise during inference.
As shown in the left illustration of \cref{fig:abla_RSR}, hole artifacts are visible on the bottom bar and edge of the table.
Our sample replacement strategy effectively eliminates these artifacts, as demonstrated in the right image of \cref{fig:abla_RSR}, by utilizing both the geometry and appearance of the object from the ROI NeRF. 
Quantitatively, as shown in \cref{tab:ablation2}, this technique significantly enhances ROI quality compared to inferring from Scene NeRF samples. 

\subsubsection{Depth-based Rays Filtering}
\label{sect:exp:abla:DRF}

\begin{figure}[ht]
	\centering
	\includegraphics[width=1\linewidth]{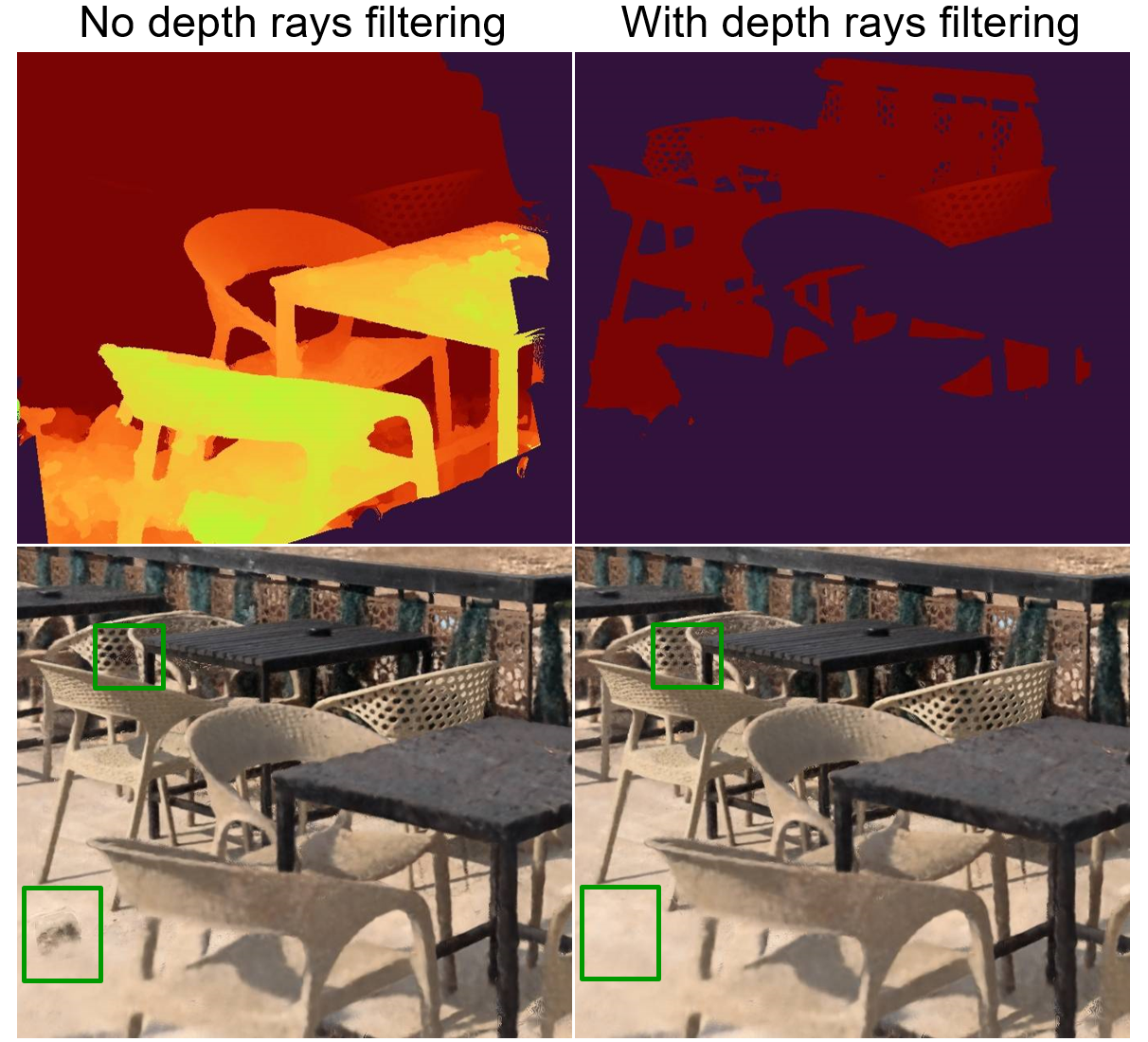}
	\caption[]{We show the rays passing through the AABB that require compositing before and after applying the depth rays filter. 
 The heatmaps illustrate a significant reduction in the number of processed rays, focusing only on those interacting with the object of interest. 
 Floaters and blurred regions, marked by the green boxes for better comparison, appear due to the unnecessary rays compositing. 
 In this case, these artifacts  formed in the empty regions of the NeRF ROI where the ROI training images are less visible and lack overall scene context to aid optimization.
    }
	\label{fig:abla_DRF}
\end{figure}

In \cref{fig:abla_DRF}, the heatmap on the left shows rays with at least one sample falling within the ROI box. 
It can be seen that the number of rays in empty spaces or occluded regions -- which do not contribute to the improvement of the ROI quality -- is significantly greater than the number of rays that interact with important geometry within the ROIs, as shown in the filtered image on the right.
This filtering significantly reduces unnecessary rays, effectively optimizing the number of rays required for compositing and thereby saving computational costs.
This helps minimize artifacts and slightly improve the inference, as seen in the rendered images.

As shown in \cref{tab:ablation2}, in \textit{Egypt}, the DRF demonstrates improvement over the baseline Full NeRF-big (2) + ROI samples (c), where the scene background is better learned with more training images. 
However, when compared to the baseline Scene NeRF-big (1) + ROI samples (c), reducing the number of composed rays limits the areas that can benefit from the ROI NeRF. 
The ROI NeRF not only learns the content inside the AABB but also predictably learns details in the proximity of the ROI. 
This makes sense since most object-focused images also include these surrounding areas. 
Consider the heatmap before filtering shown in \cref{fig:abla_DRF}, the combined rays include parts of the foreground object, $\mathit{Table1}$, which obscure the camera's view to the back target object, $\mathit{Table2}$. 
The ROI NeRF likely encodes substantial detail of these types of occluding objects, occasionally surpassing the Scene NeRF in such regions. 
As a result, a smaller number of artifacts are handled relative to the large number of rays in the surrounding region, leading to a decrease in scores when this filter is applied.
Nonetheless, we prioritize stability, reduced artifacts, and decreased computational costs since the primary goal is to enhance the ROIs while maintaining a standard LOD for the rest of the scene.
Additionally, assigning a depth to each ray is crucial for enabling multi-object composition. 
Without this DRF technique, processing and combining multiple ROIs along the same ray would become ambiguous, complex, and computationally expensive.
Indeed, as shown in \cref{tab:ablation2}, version (c) without DRF technique could not be implemented efficiently for multiple ROIs composition, so there will be no score in these cases.

\section{Limitations and Perspectives}
\label{sect:limitation}

Despite its consistent good results, our model has some limitations.
In the \textit{Hôtel de la Marine} dataset, combining Canon and iPhone shots introduces brightness and color differences. 
As NeRFs are trained separately, ROIs that are less visible to the overall view can be influenced by the close-up camera's exposure. 
Even though we calibrated the object's color using the appearance embedding feature built into the Nerfacto method, the close-up camera's exposure often dominates. 
Therefore, when compositing these objects, color shifting may appear at ROI borders.
Solutions include using a single camera type with varied lenses during data collection, applying image color transfer algorithms, or quickly refining ROI NeRF models for better alignment with the Scene NeRF.

Our Nerfacto-based composition method struggles with glossy reflective objects. 
These objects create ambiguities during training, creating virtual content mirroring nearby regions, noise clouds around reflective surfaces, and incorrect geometry interpretation.
Since our method relies heavily on accurate object geometry, particularly depth maps, this results in composition failures for reflective objects.
A simple solution is to use pixel-level compositional rendering, replacing Scene pixels within the box with ROI pixels, but this is computationally inefficient. 
The core issue lies in the representation models, which fail to accurately represent reflective surfaces. 
Integrating ideas from Ref-NeRF~\cite{verbin2022refnerf} and GSDR~\cite{ye2024gsdr} could mitigate these limitations.

The proposed approach can be naturally generalized to other NVS methods. In particular, we plan to adapt our ROI framework to \textit{3DGS}~\cite{kerbl3Dgaussians} methods to benefit from their real-time rendering capability, which allows for better and smoother scene immersion in VR.
Please refer to our supplementary material for additional  illustrations and perspectives.

\section{Conclusion}
\label{sect:conclusion}

We have presented a new framework for virtually exploring scenes, offering an improved level of detail on regions of interest, for which detailed input images are provided.
ROI-NeRFs involve multiple NeRFs that generate images for general views as well as detailed views of specific objects.
This approach takes advantage of the fast training capabilities of recent state-of-the-art methods and successfully achieves high rendering quality.
Our method not only enhances ROIs quality in the dataset but also allows for further improvement if additional images are captured and added later, without the need to retrain the entire Scene NeRF.
This approach opens up the possibility of editing and interaction with objects in the scene. 
ROI-NeRFs is a flexible downstream technique compatible with various representations, including NeRF and explicit 3D Gaussian Splatting, allowing rapid adaptation to future State-of-the-art methods.\\

\noindent 
\textbf{Data and Acknowledgments.}\\
We would like to thank the Centre des Monuments Nationaux and the team of the Hôtel de la Marine for giving us access to this magnificent place. \\
If the article is accepted, both the code and the dataset of the virtual model of the Hôtel de la Marine will be released publicly. 

\bibliographystyle{IEEEtran}
\bibliography{bibsample}

\newpage
~\newpage

\appendix


\title{Supplementary Material for ROI-NeRFs: \\Hi-Fi Visualization of Objects of Interest \\within a Scene by NeRFs Composition}

\author[1,2]{Quoc-Anh Bui}
\author[1]{Gilles Rougeron}
\author[2]{Géraldine Morin}
\author[2]{Simone Gasparini}
\affil[1]{Université Paris-Saclay, CEA, List, F-91120, Palaiseau, France}
\affil[2]{Université de Toulouse, Toulouse INP -- IRIT, France}
\renewcommand\Affilfont{\itshape\small}

\markboth{Journal of \LaTeX\ Class Files,~Vol.~14, No.~8, August~2021}%
{Shell \MakeLowercase{\textit{et al.}}: A Sample Article Using IEEEtran.cls for IEEE Journals}

\IEEEpubid{0000--0000/00\$00.00~\copyright~2021 IEEE}


\paragraph*{
This Supplementary Material complements our ROI-NeRFs paper, providing additional implementation details, evaluations, and analyses. 
It also includes a discussion of potential perspectives not covered in the main paper due to space constraints.
}

\setcounter{section}{2}

\section{Method}
\label{sect_method}

\setcounter{subsection}{2}
\subsection{Scene composition}
\label{sect:method:compo}

\begin{figure}[!t]
\centering
\includegraphics[width=1\columnwidth]{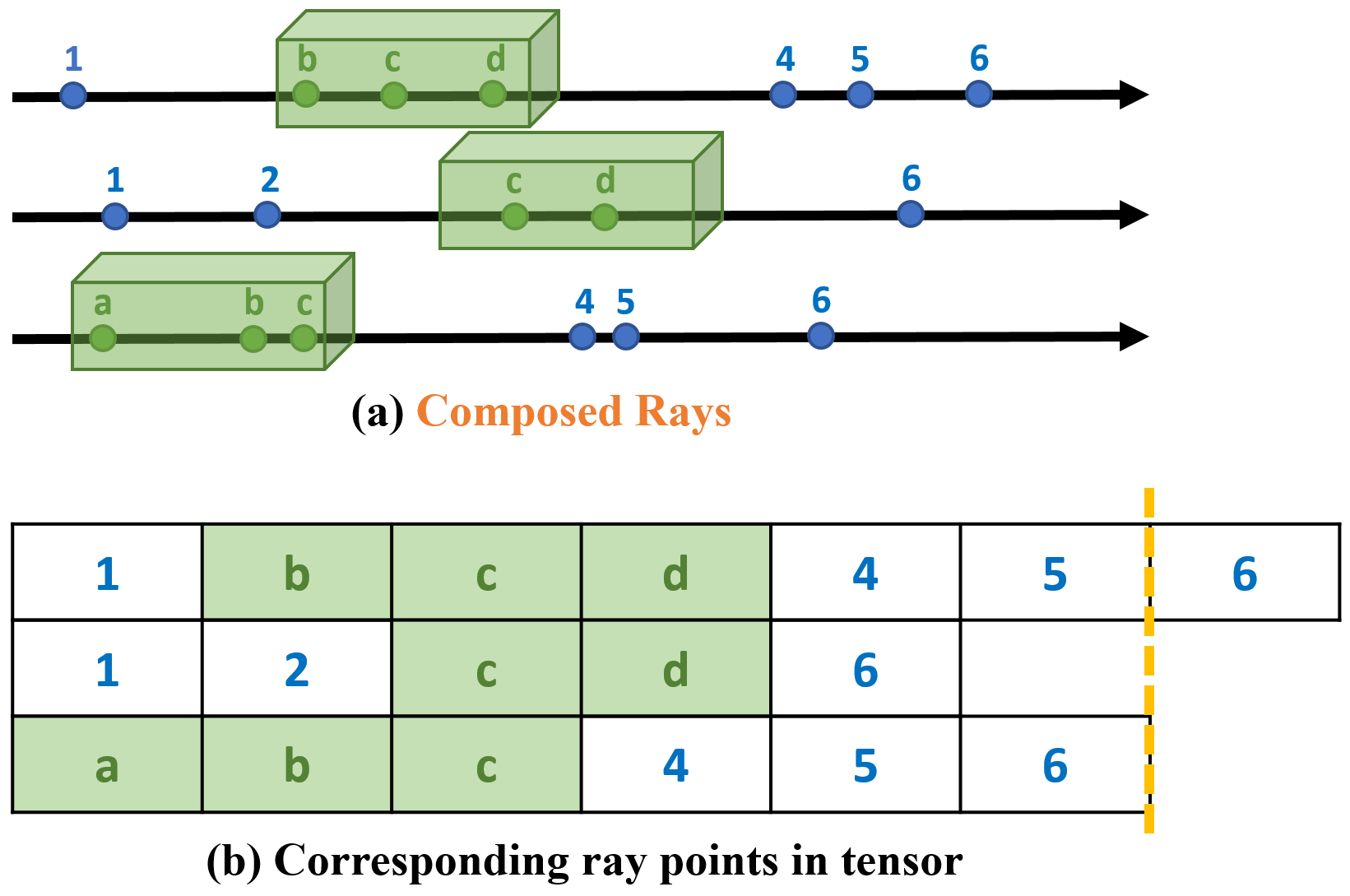}
\caption{
An example of an uneven number of points on the composed rays.
Scene samples are marked as blue with numbers, while replaced ROI points are shown as green with letters.
}
\label{fig:sample_compo}
\end{figure}

\begin{figure}[!ht]
\centering
\includegraphics[width=1\columnwidth]{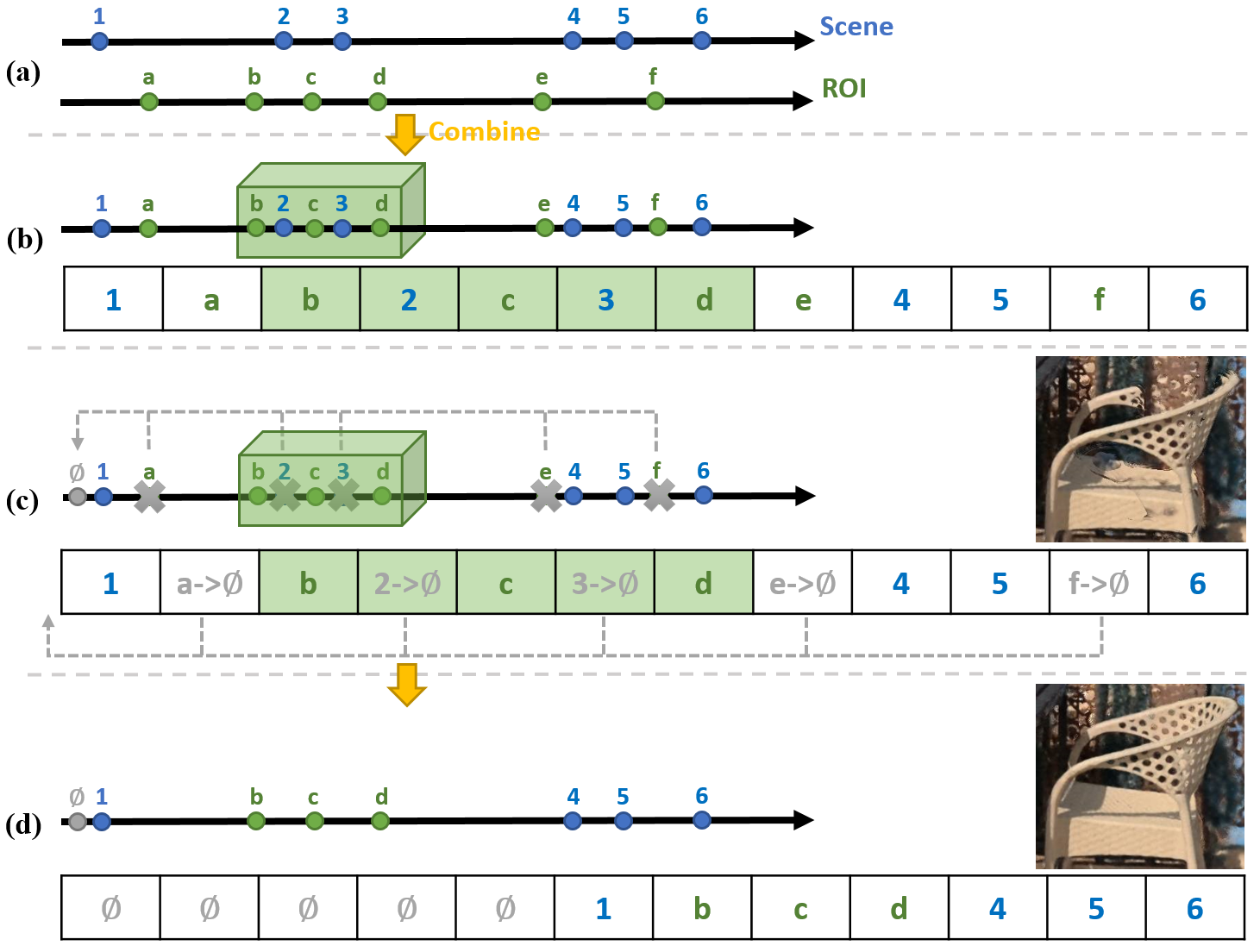}
\caption{
 Uniform composition for efficient ray aggregation.
}
\label{fig:optimize_compo}
\end{figure}

\textbf{Uniform Ray Composition.}\\
The composed rays from the ROI Samples Replacement strategy have uneven point counts due to the unpredictable number of samples within the AABB.
\cref{fig:sample_compo} shows an example of this problem. Similar to the toy example in \cref{fig:optimize_compo}, each ray from the Scene and ROI NeRFs has a fixed sample count of $6$.
In \textbf{(a)}, Scene samples are marked as blue dots with numbers, while points within the AABB replaced by the ROI are shown as green dots with letters. In \textbf{(b)}, the three tensor lines represent corresponding composed ray points with varying numbers of samples, preventing parallel tensor operations.
This mismatch prevents parallel ray processing using tensor operations.
Therefore, a naive approach of separating and processing these composed rays individually is inefficient, as it compromises the speed and efficiency of GPU parallelism, resulting in a significant and unacceptable increase in image inference time.

To address this challenge, that is, to ensure a fixed number of points on all composed rays, we propose a uniformity-preserving technique that combines all samples from both Scene and ROI rays, then making unnecessary points ``invisible'', as shown in \cref{fig:optimize_compo}.
In \textbf{(a)} and \textbf{(b)}, all samples from both Scene and ROI rays are combined to ensure an equal number of points on the composed rays, totaling $6 + 6 = 12$. 
Unnecessary points in \textbf{(c)}, such as Scene points inside the AABB (\textit{\textcolor{blue}{2}, \textcolor{blue}{3}}) and ROI points outside the AABB (\textit{\textcolor{ForestGreen}{a}, \textcolor{ForestGreen}{e}, \textcolor{ForestGreen}{f}}), are turned into invisible.
These invisible points are then set to the position of the ray origin ensuring no impact on ray aggregation, as shown in \textbf{(d)}.
These points are assigned a density of zero  and repositioned to the ray’s origin, where content is rarely found. 
This approach prevents the impact of unnecessary points, resulting in a similarity to color pixel synthesis with only valid points on rays. 
\IEEEpubidadjcol
The repositioning of empty points is crucial, as demonstrated in renders of \cref{fig:optimize_compo}, where small, thin surfaces, such as a chair's mesh, might be affected by invisibility during composition.
Furthermore, we calculate density and color only for samples contributing to ray aggregation, keeping MLP queries efficient. 
The final rendering thus reduces artifacts and maintains performance, offering a high-quality result with minimal computational overhead. 

\section{Experiments}
\label{sect:exp}

\setcounter{subsection}{1}
\subsection{Quantitative Evaluation} 
\label{sect:exp:quanti_comparison}

\begin{table*}[ht!] 
    \centering
    \caption{PSNR scores for baseline methods. In grey, cases where adding views or increasing the model size does not improve the inferred image quality.
    }
    \label{tab:baselines}
    \resizebox{0.82\textwidth}{!}{
        \begin{tabular}{l|*{4}{c}|*{6}{c}}
        Dataset & \multicolumn{4}{c|}{Egypt} & \multicolumn{6}{c}{Hôtel de la Marine} \\ 
        Methods$\vert$ROIs
        & {Table1} & {Table2} & {Table3} & {Wreath} 
        & {Carafe} & {Carving} & {Chair} & {Fruit} & {Lamp} & {Plate} \\

        \hline
        Scene NeRF-big (1)      & 20.40 & 21.33 & 22.81 & 22.20
                                & 17.71 & 21.50 & 23.39 & \fcolorbox{red}{white}{20.27} & 19.63 & 19.75 \\
        Full NeRF-big (2)       & 21.67 & 22.39 & \fcolorbox{red}{white}{23.23} & \fcolorbox{red}{white}{23.02}
                                & 17.65 & 21.37 & {23.27} & \fcolorbox{orange}{white}{20.06} & 19.55 & 19.68 \\
        \hline
        Scene NeRF-huge (3)     & 20.84 & 21.63 & 23.12 & 22.48
                                & 17.72 & 21.59 & 23.61 & \fcolorbox{red}{white}{20.31} & 19.73 & 19.87 \\
        Full NeRF-huge (4)      & 21.75 & 22.43 & \fcolorbox{orange}{white}{23.22} & \fcolorbox{orange}{white}{22.61}
                                & 17.60 & 21.47 & 23.61 & \fcolorbox{orange}{white}{20.17} & 19.65 & 19.70 \\
        \end{tabular}
    }
    
\end{table*}
\cref{tab:baselines} reports the PSNR scores for four baseline methods evaluated on the test set of each ROI across both datasets.
Scene NeRFs and Full NeRFs were trained with the default $100$k iterations to ensure optimal learning of scene details. 
However, for these Nerfacto-based methods, increasing the model size or the number of ROI images does not consistently improve rendering quality.
Indeed, for certain evaluations, such as the $\mathit{Table3}$ and $\mathit{Wreath}$ in the \textit{Egypt} dataset, the ``huge'' model showed a slight decrease in quality compared to the smaller ``big'' version. 
Similarly, in the \textit{Hôtel de la Marine} dataset, despite a larger training set with more ROI images, the Full model's PSNR score remained lower than that of the Scene model.
Thus, training the entire scene with a larger model to improve local detail quality can be both costly and inconsistent.

\begin{table*}[ht!] 
    \centering
    \caption{Quantitative evaluation of our method compared to baselines (1) to (4). The evaluations were conducted on a single ROI in both datasets. With the observation that our composition method consistently outperforms the baseline methods.}
    \label{tab:quanti_single}
    \resizebox{\textwidth}{!}{
        \begin{tabular}{l|*{6}{c}|*{6}{c}}
        Dataset & \multicolumn{6}{c|}{Egypt: Table1} & \multicolumn{6}{c}{Hôtel de la Marine: Fruit} \\ 
        Methods$\vert$Metrics
        & {PSNR\textsuperscript{$\uparrow$}} & {SSIM\textsuperscript{$\uparrow$}} & {LPIPS\textsuperscript{$\downarrow$}}& {Train} & {Render} & {Mem}
        & {PSNR\textsuperscript{$\uparrow$}} & {SSIM\textsuperscript{$\uparrow$}} & {LPIPS\textsuperscript{$\downarrow$}}& {Train} & {Render} & {Mem}\\
        \hline
        \hline
        ROI NeRF (0)                & 19.15 & 0.624 & 0.379 & 20m & 2.21s & 180MB      
                                                                                & 20.83 & 0.673 & 0.391 & 26m & 0.87s & 173MB \\
        \hline
        \hline
        Scene NeRF-big (1)      & 20.40 & 0.661 & 0.358 & 2h13m & 4.72s & 546MB        
                                                                                & 20.27 & 0.661 & 0.418 & 3h26m & 2.52s & 547MB \\
        Ours single: (1) + (0)                & \cellcolor{yellow!50}{20.64} & \cellcolor{yellow!50}{0.671} & \cellcolor{yellow!50}{0.336} & -  & 5.88s & -      
                                                                                & \cellcolor{yellow!50}{20.40} & \cellcolor{yellow!50}{0.666} & \cellcolor{yellow!50}{0.411} & -  & 3.03s & - \\
        \hline
        Full NeRF-big (2)       & 21.67 & 0.700 & 0.314 & 2h14m  & 4.70s & 546MB       
                                                                                & 20.06 & 0.650 & 0.429 & 3h46m & 2.52s & 547MB \\
        Ours single: (2) + (0)                 & \cellcolor{yellow!50}{21.78} & \cellcolor{yellow!50}{0.704} & \cellcolor{yellow!50}{0.303} & -  & 5.87s & -      
                                                                                & \cellcolor{yellow!50}{20.23} & \cellcolor{yellow!50}{0.657} & \cellcolor{yellow!50}{0.419} & -  & 3.02s & - \\
        \hline
        \hline
        Scene NeRF-huge (3)     & 20.84 & 0.688 & 0.306 & 2h49m & 4.45s & 585MB        
                                                                                & 20.31 & 0.671 & 0.366 & 4h5m & 2.37s & 586MB \\
        Ours single: (3) + (0)                 & \cellcolor{yellow!50}{21.04} & \cellcolor{yellow!50}{0.697} & \cellcolor{yellow!50}{0.293} & -  & 5.50s & -      
                                                                                & \cellcolor{yellow!50}{20.44} & \cellcolor{yellow!50}{0.672} & \cellcolor{yellow!50}{0.365} & -  & 2.79s & - \\
        \hline
        Full NeRF-huge (4)      & 21.75 & 0.721 & 0.273 & 2h50m & 4.43s & 585MB         
                                                                                & 20.17 & 0.654 & 0.401 & 4h27m & 2.37s & 586MB \\
        Ours single: (4) + (0)                 & \cellcolor{yellow!50}{21.93} & \cellcolor{yellow!50}{0.722} & \cellcolor{yellow!50}{0.270} & -  & 5.49s & -      
                                                                                & \cellcolor{yellow!50}{20.34} & \cellcolor{yellow!50}{0.660} & \cellcolor{yellow!50}{0.419} & -  & 2.79s & - \\
        \end{tabular}
    }
\end{table*}
\cref{tab:quanti_single} presents the quantitative evaluations of our proposed method.
We compare the results of each baseline version ($1$--$4$) with our compositional approach, which combines the ROI NeRF with the baseline models. 
The experiments focus on a single ROI per dataset, $\mathit{Table1}$ in \textit{Egypt} and $\mathit{Fruit}$ in \textit{Hôtel de la Marine}, to avoid any bias that a varying number of ROIs could introduce in the views.

\begin{figure*}[ht!]
	\centering
	\includegraphics[width=1\linewidth]{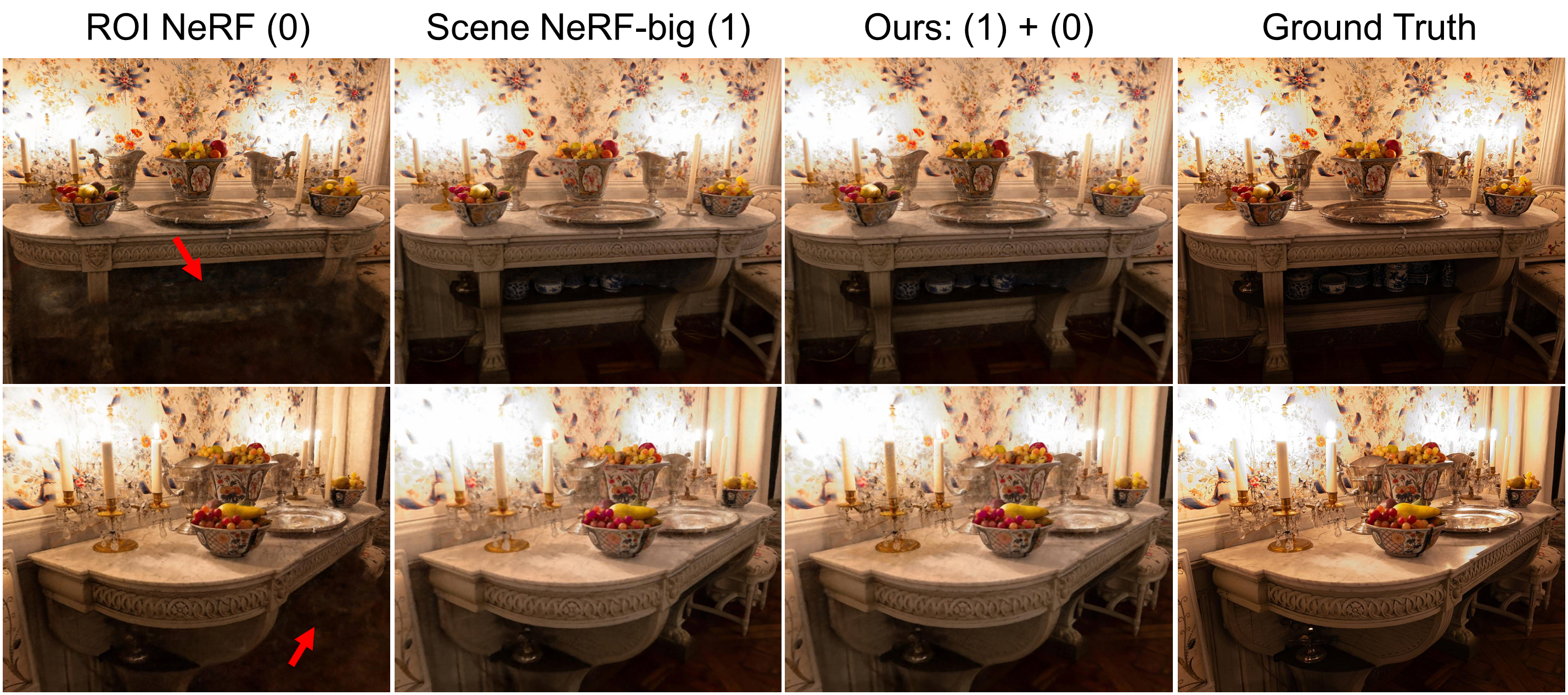}
	\caption[]{Missing geometric information in general views of the ROI NeRFs, compared to the Scene NeRF method and our approach, indicated by red arrows.
 }
	\label{fig:hotel_roi_artifact}
\end{figure*}

As noted, since ROI NeRFs lacked information about the surrounding scene, this led to worse scores compared to baseline methods on test sets with sufficiently distant cameras, where the ROI’s occupied area in the image is not too big.
\cref{tab:quanti_single} confirms this for the \textit{Egypt} dataset.
However, in the \textit{Hôtel de la Marine} dataset, the ROI NeRF scores are better than the baselines. 
This can be explained by the nature of the dataset, which consists mainly of two types of camera views: (1) close-up views of the table area containing all the objects of interest, where the objects occupy a large part of the image, and (2) wider views, which are excluded from the ROI set during camera selection due to their distance.
In addition, the large number of close-up images selected for training the ROI NeRFs results in the neighboring regions also being encoded in detail. 
Taken together, these factors result in significantly better ROI scores than the baselines. 
However, despite the better scores driven by the dominance of ROIs in the rendered image, artifacts and missing geometric information persist due to the limited generality of the ROI NeRF, as illustrated in \cref{fig:hotel_roi_artifact}.

\begin{table*}[ht!] 
    \centering
    \caption{
    PSNR scores for our composition methods on objects of the Egypt dataset.
    }
    \label{tab:details_compo}
    \resizebox{\textwidth}{!}{
        \begin{tabular}{l|*{5}{c}|*{7}{c}|}
        Dataset & \multicolumn{5}{c|}{Egypt} & \multicolumn{7}{c|}{Hôtel de la Marine} \\ 
        Methods$\vert$ROIs
        & {Table1} & {Table2} & {Table3} & {Wreath} & \multicolumn{1}{|c|}{Mean}
        & {Carafe} & {Carving} & {Chair} & {Fruit} & {Lamp} & {Plate} & \multicolumn{1}{|c|}{Mean} \\

        \hline
        Scene NeRF-big (1)  & 20.40 & 21.33 & 22.81 & 22.20 & \multicolumn{1}{|c|}{21.69}
                            & 17.71 & 21.50 & 23.39 & 20.27 & 19.63 & 19.75 & \multicolumn{1}{|c|}{20.37} \\
        Ours single (1)         & \cellcolor{yellow!50}{20.64} & \cellcolor{yellow!50}{21.60} & \cellcolor{yellow!50}{22.94} & \cellcolor{yellow!50}{22.25} & \multicolumn{1}{|c|}{\cellcolor{yellow!50}{21.86}}
                                & \cellcolor{yellow!50}{17.77} & \cellcolor{yellow!50}{21.52} & \cellcolor{yellow!50}{23.55} & \cellcolor{yellow!50}{20.40} & \cellcolor{yellow!50}{19.72} & \cellcolor{yellow!50}{19.76} & \multicolumn{1}{|c|}{\cellcolor{yellow!50}{20.45}} \\
        Ours multiple (1)   & \cellcolor{orange!50}{20.76} & \cellcolor{orange!50}{21.92} & \cellcolor{orange!50}{23.21} & \cellcolor{orange!50}{22.35} & \multicolumn{1}{|c|}{\cellcolor{orange!50}{22.06}}
                            & \cellcolor{orange!50}{18.11} & \cellcolor{orange!50}{21.55} & \cellcolor{orange!50}{23.49} & \cellcolor{orange!50}{20.46} & \cellcolor{orange!50}{19.91} & \cellcolor{orange!50}{20.02} & \multicolumn{1}{|c|}{\cellcolor{orange!50}{20.59}} \\
        \hline
        Full NeRF-big (2)   & 21.67 & 22.39 & 23.23 & 23.02 & \multicolumn{1}{|c|}{22.58}
                            & 17.65 & 21.37 & 23.27 & 20.06 & 19.55 & 19.68 & \multicolumn{1}{|c|}{20.26} \\
        Ours single (2)         & \cellcolor{yellow!50}{21.78} & \cellcolor{yellow!50}{22.41} & \cellcolor{yellow!50}{23.35} & \cellcolor{yellow!50}{23.05} & \multicolumn{1}{|c|}{\cellcolor{yellow!50}{22.65}}
                                & \cellcolor{yellow!50}{17.77} & \cellcolor{yellow!50}{21.40} & \cellcolor{yellow!50}{23.44} & \cellcolor{yellow!50}{20.23} & \cellcolor{yellow!50}{19.64} & \cellcolor{yellow!50}{19.69} & \multicolumn{1}{|c|}{\cellcolor{yellow!50}{20.36}} \\
        Ours multiple (2)   & \cellcolor{orange!50}{21.86} & \cellcolor{orange!50}{22.62} & \cellcolor{orange!50}{23.60} & \cellcolor{orange!50}{23.13} & \multicolumn{1}{|c|}{\cellcolor{orange!50}{22.80}}
                            & \cellcolor{orange!50}{18.13} & \cellcolor{orange!50}{21.45} & \cellcolor{orange!50}{23.42} & \cellcolor{orange!50}{20.33} & \cellcolor{orange!50}{19.87} & \cellcolor{orange!50}{20.01} & \multicolumn{1}{|c|}{\cellcolor{orange!50}{20.53}} \\
        \hline
        Scene NeRF-huge (3) & 20.84 & 21.63 & 23.12 & 22.48 & \multicolumn{1}{|c|}{22.02}
                            & 17.72 & 21.59 & 23.61 & 20.31 & 19.73 & 19.87 & \multicolumn{1}{|c|}{20.47} \\
        Ours single (3)         & \cellcolor{yellow!50}{21.04} & \cellcolor{yellow!50}{21.93} & \cellcolor{yellow!50}{23.19} & \cellcolor{yellow!50}{22.51} & \multicolumn{1}{|c|}{\cellcolor{yellow!50}{22.17}}
                                & \cellcolor{yellow!50}{17.79} & \cellcolor{yellow!50}{22.00} & \cellcolor{yellow!50}{23.63} & \cellcolor{yellow!50}{20.44} & \cellcolor{yellow!50}{19.82} & \cellcolor{yellow!50}{19.87} & \multicolumn{1}{|c|}{\cellcolor{yellow!50}{20.54}} \\
        Ours multiple (3)   & \cellcolor{orange!50}{21.11} & \cellcolor{orange!50}{22.17} & \cellcolor{orange!50}{23.41} & \cellcolor{orange!50}{22.58} & \multicolumn{1}{|c|}{\cellcolor{orange!50}{22.32}}
                            & \cellcolor{orange!50}{18.19} & \cellcolor{orange!50}{22.01} & \cellcolor{orange!50}{23.75} & \cellcolor{orange!50}{20.53} & \cellcolor{orange!50}{20.00} & \cellcolor{orange!50}{20.14} & \multicolumn{1}{|c|}{\cellcolor{orange!50}{20.68}} \\
        \hline
        Full NeRF-huge (4)  & 21.75 & 22.43 & 23.22 & 22.61 & \multicolumn{1}{|c|}{22.50}
                            & 17.60 & 21.47 & 23.61 & 20.17 & 19.65 & 19.70 & \multicolumn{1}{|c|}{20.37} \\
        Ours single (4)         & \cellcolor{yellow!50}{21.93} & \cellcolor{yellow!50}{22.42} & \cellcolor{yellow!50}{23.31} & \cellcolor{yellow!50}{22.63} & \multicolumn{1}{|c|}{\cellcolor{yellow!50}{22.57}}
                                & \cellcolor{yellow!50}{17.72} & \cellcolor{yellow!50}{21.48} & \cellcolor{yellow!50}{23.66} & \cellcolor{yellow!50}{20.34} & \cellcolor{yellow!50}{19.74} & \cellcolor{yellow!50}{19.74} & \multicolumn{1}{|c|}{\cellcolor{yellow!50}{20.46}} \\
        Ours multiple (4)   & \cellcolor{orange!50}{21.99} & \cellcolor{orange!50}{22.62} & \cellcolor{orange!50}{23.53} & \cellcolor{orange!50}{22.69} & \multicolumn{1}{|c|}{\cellcolor{orange!50}{22.71}}
                            & \cellcolor{orange!50}{18.18} & \cellcolor{orange!50}{21.50} & \cellcolor{orange!50}{23.73} & \cellcolor{orange!50}{20.46} & \cellcolor{orange!50}{19.99} & \cellcolor{orange!50}{20.06} & \multicolumn{1}{|c|}{\cellcolor{orange!50}{20.64}} \\
        \end{tabular}
    }
\end{table*}
\begin{table*}[ht!] 
    \centering
    \caption{Range of improvement of our single-composition method over baselines on each test image of a single ROI in both datasets.}
    \label{tab:interval}
    \resizebox{0.9\textwidth}{!}{
        \begin{tabular}{l|*{3}{c}|*{3}{c}}
        Dataset & \multicolumn{3}{c|}{Egypt: Table1} & \multicolumn{3}{c}{Hôtel de la Marine: Fruit} \\ 
        Baselines$\vert$ Improvement Range
        & {PSNR\textsuperscript{$\uparrow$}} & {SSIM\textsuperscript{$\uparrow$}} & {LPIPS\textsuperscript{$\downarrow$}} 
        & {PSNR\textsuperscript{$\uparrow$}} & {SSIM\textsuperscript{$\uparrow$}} & {LPIPS\textsuperscript{$\downarrow$}} \\

        \hline
        Scene NeRF-big (1)      & [0, 0.51] & [0, 0.024] & [0, 0.043]        
                                & [0.01, 0.78] & [0.003, 0.027] & [0.003, 0.042] \\
        Full NeRF-big (2)       & [0, 0.57] & [0, 0.018] & [0, 0.029]        
                                & [0.01, \cellcolor{yellow!50}{1.00}] & [0.003, 0.033] & [0.002, 0.051] \\
        \hline
        Scene NeRF-huge (3)     & [0, 0.52] & [0, 0.023] & [0, 0.028]        
                                & [0.04, 0.75] & [0.004, 0.008] & [0.004, 0.005] \\
        Full NeRF-huge (4)      & [0, 0.75] & [0, 0.012] & [0, 0.013]        
                                & [0.06, \cellcolor{yellow!50}{1.10}] & [0.002, 0.023] & [0.003, 0.031] \\
        \end{tabular}
    }
\end{table*}
\cref{tab:details_compo} provide detailed PSNR scores for our method, including the multiple-composition approach, for each object in both datasets. 
The multiple-composition method combines all valid visible ROIs in the corresponding inference view, using the same test images as the single-composition and baseline methods to ensure compatibility and fairness.
Similarly, our method of compositing multiple ROIs also consistently improves the rendering quality of objects for any model and outperforms the single-composition method.
This is reasonable because the quality improvement depends on the number of ROIs visible in the render view.

\begin{table}[ht!] 
    \centering
    \caption{
    Average PSNR scores within ROI boxes for our single-composition method compared to baseline methods.
    }
    \label{tab:aabb}
    \resizebox{0.4\textwidth}{!}{
        \begin{tabular}{l|*{1}{c}|*{1}{c}}
        Methods$\vert$Dataset
        & {Egypt} & {Hôtel de la Marine} \\
        \hline
        \hline
        ROI NeRF (0)            & 22.27 & 21.36 \\
        
        \hline
        \hline
        Scene NeRF-big (1)      & 21.12 & 19.85 \\
        Ours single: (1) + (0)                    & \cellcolor{yellow!50}{21.51} & \cellcolor{yellow!50}{20.39} \\
        \hline
        Full NeRF-big (2)       & 22.07 & 19.73 \\
        Ours single: (2) + (0)                    & \cellcolor{yellow!50}{22.25} & \cellcolor{yellow!50}{20.43} \\
        \hline
        \hline
        Scene NeRF-huge (3)     & 21.50 & 19.93 \\
        Ours single: (3) + (0)                    & \cellcolor{yellow!50}{21.81} & \cellcolor{yellow!50}{20.45} \\
        \hline
        Full NeRF-huge (4)      & 22.05 & 20.00 \\
        Ours single: (4) + (0)                    & \cellcolor{yellow!50}{22.19} & \cellcolor{yellow!50}{20.61} \\
        \end{tabular}
    }
    
\end{table}
We also evaluated the single-composition methods only in the ROI's AABBs, as shown in \cref{tab:aabb}.
The improvement in the ROI of our method over the baselines is substantial, particularly when contrasted with the corresponding scores calculated over the whole image in \cref{tab:quanti_single}, which are drowned out by the other part of the images.
Additionally, the PSNR scores for ROI NeRFs within the AABB are higher than those for the whole image, which demonstrates that the LOD is only high in the ROI.

\setcounter{subsection}{3}
\subsection{Ablations} 
We provide a full evaluation of our method's components' impacts on all four baseline methods in Ablation \cref{tab:ablation}.
\begin{table}[h] 
    \centering
    \caption{PSNR scores for Ablations. This study is conducted on all four baseline methods and their compositions with single or multiple ROIs across all ROIs in both datasets. The experiments include: (a) No ROI Samples Replacement (RSR) and no Depth Rays Filtering (DRF), using only Scene samples for composition, (b) No RSR, with DRF only, (c) RSR without DRF, and (d) our complete model.
    }
    \label{tab:ablation}
    \resizebox{0.5\textwidth}{!}{
        \begin{tabular}{l|*{1}{c}|*{1}{c}|*{1}{c}|*{1}{c}}
        Dataset & \multicolumn{2}{c|}{Egypt} & \multicolumn{2}{c}{Hôtel de la Marine} \\ 
        Methods$\vert$Mode
        & {Single} & {Multiple} & {Single} & {Multiple} \\
        \hline
        Scene NeRF-big (1)      & \multicolumn{2}{c|}{21.69} & \multicolumn{2}{c}{20.37}\\
        \hline
        No RSR, No DRF (a)      & 21.73 & 21.76 & 20.43 & 20.57 \\
        No RSR (b)              & 21.71 & 21.73 & 20.43 & 20.54 \\
        No DRF (c)              & \cellcolor{orange!50}{21.89} & /     & \cellcolor{yellow!50}{20.44} & / \\
        Full (d)                & \cellcolor{yellow!50}{21.86} & \cellcolor{orange!50}{22.06} & \cellcolor{orange!50}{20.45} & \cellcolor{orange!50}{20.59} \\

        \hline
        Full NeRF-big (2)       & \multicolumn{2}{c|}{22.58} & \multicolumn{2}{c}{20.26}\\
        \hline
        No RSR, No DRF (a)      & 22.46 & 22.39 & 20.36 & 20.51 \\
        No RSR (b)              & 22.46 & 22.39 & 20.35 & 20.48 \\
        No DRF (c)              & \cellcolor{yellow!50}{22.64} & /     & \cellcolor{yellow!50}{20.36} & / \\
        Full (d)                & \cellcolor{orange!50}{22.65} & \cellcolor{orange!50}{22.80} & \cellcolor{orange!50}{20.37} & \cellcolor{orange!50}{20.53} \\

        \hline
        Scene NeRF-huge (3)     & \multicolumn{2}{c|}{22.02} & \multicolumn{2}{c}{20.47}\\
        \hline
        No RSR, No DRF (a)      & 21.96 & 21.86 & 20.53 & 20.58 \\
        No RSR (b)              & 21.95 & 21.86 & 20.52 & 20.56 \\
        No DRF (c)              & \cellcolor{orange!50}{22.19} & /     & \cellcolor{yellow!50}{20.54} & / \\
        Full (d)                & \cellcolor{yellow!50}{22.17} & \cellcolor{orange!50}{22.32} & \cellcolor{orange!50}{20.55 }& \cellcolor{orange!50}{20.68} \\

        \hline
        Full NeRF-huge (4)      & \multicolumn{2}{c|}{22.50} & \multicolumn{2}{c}{20.37}\\
        \hline
        No RSR, No DRF (a)      & 22.35 & 22.23 & 20.45 & 20.57 \\
        No RSR (b)              & 22.36 & 22.24 & 20.44 & 20.55 \\
        No DRF (c)              & \cellcolor{yellow!50}{22.55} & /     & \cellcolor{yellow!50}{20.46} & / \\
        Full (d)                & \cellcolor{orange!50}{22.57} & \cellcolor{orange!50}{22.71} & \cellcolor{orange!50}{20.47} & \cellcolor{orange!50}{20.64} \\

        \end{tabular}
    }
\end{table}

\section{Limitations and Perspectives}
\label{sect:limitation}

\begin{figure}[ht]
	\centering
	\includegraphics[width=1\linewidth]{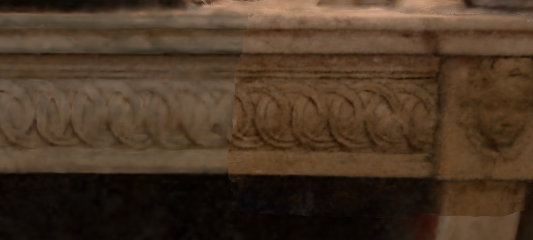}
	\caption[]{The color shifting issue occurs when compositing an ROI NeRF trained on close-up images that differ in color and brightness from those used for the Scene NeRF. In this example, while the pattern details on the stone table are enhanced, the carved area shows a color mismatch with the background, causing the ROI to stand out from the rest of the image.
 }
	\label{fig:limit_color_shift}
\end{figure}

\begin{figure}[ht]
	\centering
	\includegraphics[width=1\linewidth]{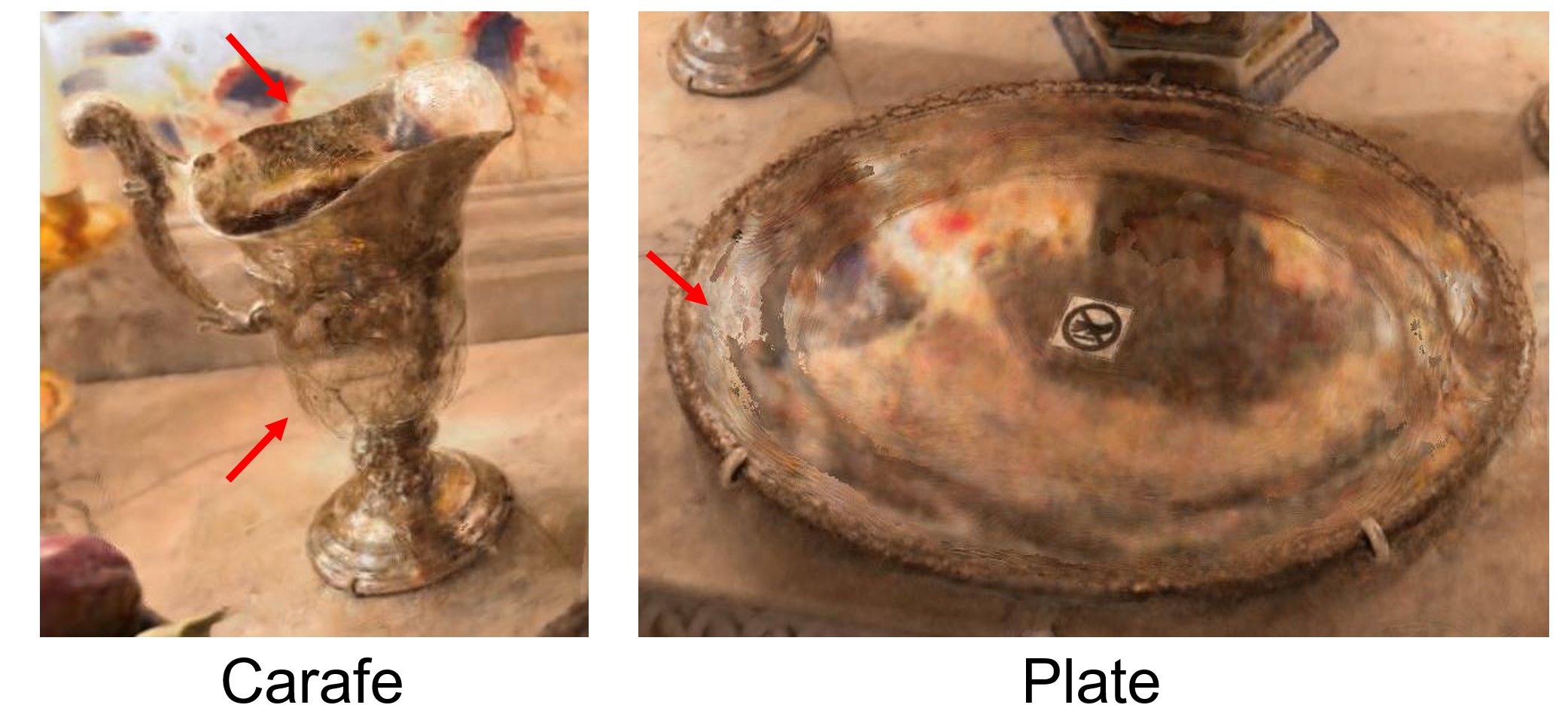}
	\caption[]{
     Failure cases in composing glossy reflective objects. Inferred images from the proposed composition method are shown for two objects in the \textit{Hôtel de la Marine} dataset, with noisy composition patches occurring in ROIs highlighted by red arrows.
 }
	\label{fig:limit_glossy}
\end{figure}

As noted about the limitations of our methods, we provide example images of artifacts in the ROIs during compositing. \cref{fig:limit_color_shift} shows the color shifting issue, while \cref{fig:limit_glossy} illustrates the failure on reflective objects.

While ROIs may typically be implicitly defined during the image-capturing stage due to more detailed captures, in our proof-of-concept they are manually selected by the user by drawing AABBs around objects in the sparse point cloud space to simplify processing.
Alternatively, AABBs could be automatically generated based on user-selected objects or identified autonomously in regions with a high concentration of input images.
Future developments could explore methods where users localize objects in a few images using manual selection or object segmentation tools, such as the Segment Anything Model (SAM)~\cite{kirillov2023segany}.
The 3D AABBs could then be automatically generated by consistently identifying visible 3D keypoints in the 2D images or through point region classification using 3D point clustering algorithms, like HDBSCAN~\cite{mcinnes2017hdbscan}.

The Camera Selection technique for scene decomposition currently relies solely on the number of visible keypoints to select cameras.
While this approach provides stable performance for partitioning, we still lack a clear definition of how many images focusing on different parts of the ROI are needed to effectively train the ROI NeRF. 
To address this issue, we could integrate the uncertainty estimation of the information gain in scene understanding to identify the most informative views.
Thus, we could efficiently capture and add images to the training set efficiently with minimal additional resources as shown in \textit{ActiveNeRF}~\cite{pan2022activenerf} or \textit{IOVS4NeRF}~\cite{xie2024iovs4nerf}.

The scene decomposition and recomposition strategy has proven to enhance rendering quality consistently. 
A potential direction for future work is hierarchical decomposition and composition. 
For instance, at level 1, large objects like dining tables can be separated and trained at the corresponding LOD, followed by level 2, where details such as patterns or small objects on the table are trained at a finer LOD. 
This process can continue for additional levels. 
During inference, depending on the viewpoint, the NeRF of each component at each level will be selected appropriately to achieve LOD-focused high-quality rendering.
This approach could also be applied to articulated objects, inspired by \textit{NARF22}~\cite{lewis2022narf}.

An unexplored aspect of our method is the ability to edit objects in the scene. 
To implement this, it is necessary to tighten AABBs around the object or preferably to define the mask of the object of interest.
We could perform segmentation on 2D images to reconstruct consistent masks for objects of interest, inspired by \textit{ObjectNeRF}~\cite{yang2021objectnerf} and \textit{UDC-NeRF}~\cite{wang2023udcnerf}.
AABBs could be refined either by intersecting these object masks through space or by propagating image mask IDs to the 3D sparse point cloud, in a similar fashion as \textit{K3BO}~\cite{10222520}.
Another approach is to employ 3D segmentation after estimating the object's geometry through rapid training of a NeRF model on the object. 
After segmentation, objects could be removed, replaced, or moved. 
However, this creates new issues as the scene illumination has to remain consistent while object appearance is embedded in their respective NeRFs. 
Similarly, any holes left under moved objects should be filled in.
Inpainting methods like those in \textit{Removing Objects From NeRF}~\cite{Weder2023Removing} could then be applied.



\vfill

\end{document}